\documentclass[letterpaper]{article}
\usepackage[preprint]{aaai2027}
\usepackage[hyphens]{url}
\usepackage{graphicx}
\usepackage{natbib}
\usepackage{caption}
\usepackage{amsmath}
\usepackage{amssymb}
\usepackage{booktabs}
\usepackage{array}

\newtheorem{theorem}{Theorem}
\newtheorem{proposition}{Proposition}
\newtheorem{lemma}{Lemma}
\newenvironment{proofenv}{\par\noindent\textit{Proof.}\ }{\hfill$\square$\par}
\newcounter{algorithm}

\newcommand{\Qstop}{Q_{\mathrm{stop}}}
\newcommand{\Qcont}{Q_{\mathrm{continue}}}
\newcommand{\Uex}{U_{\mathrm{exact}}}
\newcommand{\Upar}{U_{\mathrm{partial}}}
\newcommand{\keymetric}[1]{{\bfseries\boldmath#1}}%

\title{Scores Are Not Decisions: Cost-Aware Stopping for Tool Acquisition in LLM Agents}
\author {
    Yicheng Feng\textsuperscript{\rm 1},
    Yan Zhang\textsuperscript{\rm 2},
    Yan Cheng\textsuperscript{\rm 3}\corresponding,
    Wei Qi\textsuperscript{\rm 4}\corresponding
}
\affiliations {
    \textsuperscript{\rm 1}School of Advanced Manufacturing and Robotics, Peking University\\
    \textsuperscript{\rm 2}Desautels Faculty of Management, McGill University\\
    \textsuperscript{\rm 3}Business School, Shanghai University of Finance and Economics\\
    \textsuperscript{\rm 4}Department of Industrial Engineering, Tsinghua University\\
    yf2388@stu.pku.edu.cn, yan.zhang13@mail.mcgill.ca, chengyan@mail.shufe.edu.cn, qiw@tsinghua.edu.cn
}

\begin{document}

\maketitle

\begin{abstract}

As LLM agents increasingly depend on diverse external services such as search engines, databases, and connectors, agent harnesses face a fundamental tool-selection challenge: acquiring too few tools leaves the task under-informed, while too many adds cost, context load, and privacy exposure. Routers and retrievers can rank candidate tools by relevance, but a ranking alone does not determine how many are worth selecting. Existing approaches leave acquisition under heterogeneous costs unaddressed. We formulate this decision as cost-aware marginal decision-focused stopping (CAM-DF) over ranked tool prefixes, with CAM-DF-lite as a compact interpretable variant. We train directly on the offline gap between stopping now and the best continuation: its sign labels the decision, its magnitude weights each error by the payoff at stake. We prove this objective is Bayes-aligned with the stopping target and that score-only rules are suboptimal under heterogeneous costs. We evaluate on 1,343 tasks across five tool-use domains. On $\tau$-bench Retail, CAM-DF attains the highest payoff among deployable methods, with gains over a predict-then-threshold baseline across all five ranking sources and two cost regimes. Our approach is state-of-the-art under heterogeneous costs and high cost pressure, with larger gains under weaker rankings. In live execution, CAM-DF exposes the agent to 37\% fewer tools than full access while maintaining comparable task success. The CAM-DF family is a lightweight pre-execution plugin that turns existing tool rankings into lower-cost acquisition decisions without fine-tuning the underlying LLM.

\end{abstract}

\section{Introduction}
\label{sec:intro}

As tool ecosystems expand, agent harnesses must decide how much external tool access to provide for each task. Unnecessary tool acquisition can consume context and system resources, add latency and API expense, and increase privacy and error exposure, with these costs accumulating across repeated agent turns. Recent work shows that performance can plateau as tool-call budgets increase \citep{liu2025budgetaware}, while even leading models do not reliably execute cost-optimal tool-use plans as tasks become more complex or tool costs change \citep{liu2025costbench}. Agent-evaluation research has accordingly argued that cost should be assessed alongside task performance \citep{kapoor2024aiagents}. This increased focus highlights a central question: when does expanded tool access justify its additional cost?

We examine this question in a general pre-execution setting in which an upstream scoring or routing mechanism supplies a task-specific ranking of candidate tools with heterogeneous costs. Given this ranking, the agent harness must still choose an acquisition depth $k$. This depth defines a prefix containing the first $k$ tools in the ranked list. The ranking establishes the relative priority of candidate tools, but it does not by itself identify whether an individual tool is necessary or whether the selected prefix contains all tools required for the task. Determining where to stop therefore requires comparing the payoff from stopping with the best payoff attainable by continuing. Moreover, when tool costs differ, score-based ranking is provably inconsistent with the optimal acquisition rule (Sec.~\ref{sec:theory}). Prior work instead routes queries among predefined retrieval strategies \citep{jeong2024adaptiverag} or learns whether to continue after each round of iterative retrieval \citep{park2025stoprag}. These formulations leave open how to convert a given tool ranking into a cost-aware acquisition depth before execution.

\begin{figure*}[t]
\centering
\includegraphics[width=0.8\textwidth]{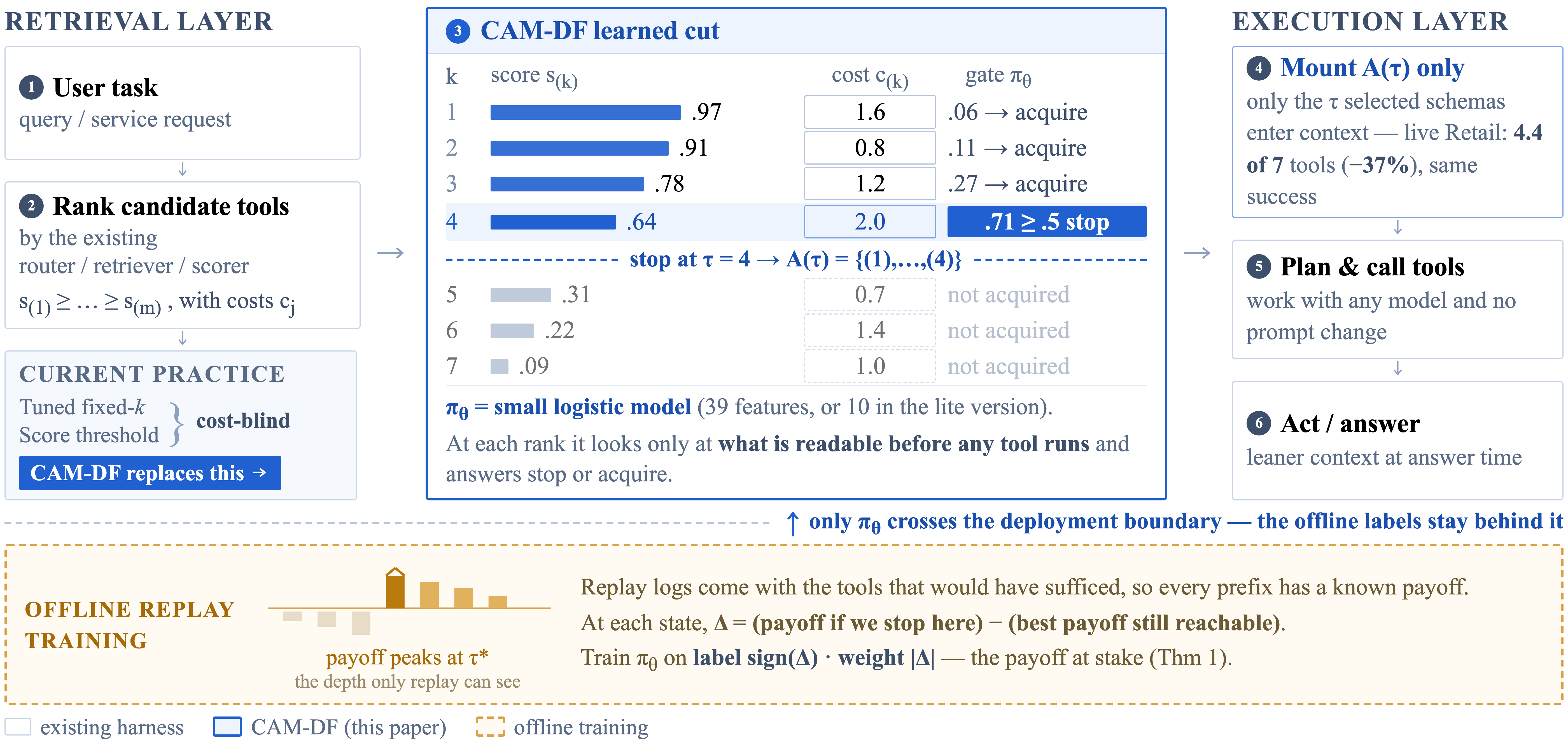}
\caption{CAM-DF as a pre-execution stop plugin inside one agent turn.
The agent flow is unchanged: a router, retriever, scoring prompt, or shadow pass exposes a candidate order.
CAM-DF virtually walks the ranking and fixes the prefix $A(\tau)$ before any selected tool executes. Offline labels define the regret-weighted target; deployment sees public features.}
\label{fig:framework}
\end{figure*}

To address this gap, we develop Cost-Aware Marginal Decision-Focused Stopping (CAM-DF), a learned stopping policy over ranked tool prefixes. CAM-DF derives supervision directly from downstream payoff. For task $x$ and acquired tool set $A$, we define $U(A,x)=\operatorname{sufficiency}(A,x)-\lambda\sum_{j\in A}c_j$, where $\operatorname{sufficiency}(A,x)$ indicates whether $A$ contains the required tools, $c_j$ denotes tool cost, and $\lambda$ controls cost pressure. For each candidate stopping point, we compare the payoff of the current prefix with the highest payoff attainable from any longer prefix. The sign of this payoff gap determines the target decision, while its magnitude weights errors by the payoff at stake. We show that the resulting regret-weighted objective is Bayes-aligned with the offline stopping target. As shown in Figure~\ref{fig:framework}, CAM-DF uses ranking scores, tool costs, and prefix-progress features to select an acquisition depth. The downstream agent then operates with the selected prefix, requiring no modification to the upstream ranking mechanism or the underlying LLM.

We evaluate CAM-DF on 1{,}343 tasks across five tool-use domains, using $\tau$-bench Retail \citep{yao2024taubench} as the primary testbed. CAM-DF achieves the highest mean payoff among deployable methods under both uniform and heterogeneous tool costs. Across five frozen LLM ranking configurations and two cost regimes, it outperforms a feature-matched predict-then-threshold baseline in all ten settings, with 95\% paired-bootstrap confidence intervals excluding zero in nine. A tuned score-per-cost threshold remains competitive at $\lambda=0.12$, but CAM-DF's advantage becomes clearer at $\lambda=0.20$; its gains over predict-then-threshold stopping are largest under the weakest ranking configuration. Ablations indicate that marginal-cost features matter most under heterogeneous costs and that learning the payoff gap is more effective than predicting current-prefix sufficiency. Its gains across four additional domains likewise concentrate under heterogeneous costs, although the tuned threshold remains competitive in some regimes. In an end-to-end $\tau$-bench Retail evaluation, CAM-DF reduces pre-execution read-tool exposure from 7 to 4.4 tools (37\%) while maintaining comparable observed task success. We further introduce CAM-DF-lite, a lightweight, interpretable variant that retains most of the full policy's empirical gains.

Our contributions are threefold. First, we formulate tool acquisition as cost-aware stopping over ranked prefixes, enabling acquisition depth to be optimized directly for downstream utility while accounting for task sufficiency and heterogeneous costs. Second, we develop a decision-focused objective with Bayes-alignment guarantees and a marginal value-per-cost analysis that characterizes when score-only acquisition becomes suboptimal. Third, we propose the CAM-DF family of deployable, modular stopping policies, comprising a flexible learned policy and a lightweight, interpretable variant that retains most of the full policy's empirical gains. Taken together, these contributions position cost-aware stopping as a well-founded and practical control layer for agent harnesses.

% The remainder of this paper is organized as follows: Section~\ref{sec:related} reviews related work; Sections~\ref{sec:problem} and~\ref{sec:theory} present the problem formulation and theoretical analysis; Section~\ref{sec:method} introduces the CAM-DF family; Section~\ref{sec:experiments} reports the multi-domain evaluation; and Section~\ref{sec:conclusion} discusses limitations and future directions.

\section{Related Work}
\label{sec:related}

CAM-DF connects research on cost-aware tool use and adaptive ranked-list depth with methodological work on decision-focused learning and costly search.

\paragraph{Tool routing and cost-aware tool use.}
Tool-augmented LLM systems support API invocation and candidate retrieval from large catalogs \citep{schick2023toolformer,patil2024gorilla,qin2024toolllm,du2024anytool}. Work on tool-use control studies tool necessity and selection, on-demand registration, and reductions in unnecessary calls or context exposure \citep{huang2024metatool,ross2025when2call,zhang2024ecoact,qian2025smart,wu2026tocall,sun2026when2tool}. Cost-aware methods further allocate tools under budgets, generate tool-use plans, or manage exploration costs during sequential execution \citep{zheng2024budgettool,wu2025catp,ding2026calibrate,liu2026utilityguided}. These studies primarily focus on tool selection, call timing, or budget allocation, without directly modeling acquisition depth after ranking and before execution. CAM-DF adds this control layer to separate tool prioritization from access allocation, enabling task-specific acquisition depth without modifying the upstream ranker or downstream agent.

\paragraph{Adaptive retrieval and ranked-list depth.}
Adaptive retrieval regulates evidence gathering using confidence, learned retrieval signals, attention-based information needs, or predicted query complexity \citep{jiang2023flare,asai2024selfrag,su2024dragin,jeong2024adaptiverag}. Within this literature, two lines are particularly relevant to CAM-DF. The first treats iterative retrieval as a sequential stopping problem and learns whether to continue after each retrieval round \citep{park2025stoprag}. The second learns query-specific cutoffs for retrieval, reranking, and ranked tool descriptions, including joint optimization of order and depth \citep{bahri2020choppy,meng2024rlt,sun2025dynamicrag,repantis2026howmanytools}. These lines establish learned stopping and query-specific depth as important control mechanisms, but their objectives do not explicitly balance downstream task sufficiency against heterogeneous acquisition costs. CAM-DF addresses this gap by incorporating both terms into downstream payoff and learning cutoffs from stop-versus-continue regret.

\paragraph{Decision-focused learning.}
Decision-focused learning incorporates downstream objectives because statistically similar prediction errors can produce different decision losses \citep{mandi2024dfl}. Related approaches learn prescriptions directly from contextual data \citep{ban2019bigdata,bertsimas2022datadriven}, differentiate through downstream optimization \citep{donti2017task}, develop regret-based surrogate losses \citep{elmachtoub2022spo}, or adjust fitted estimates for asymmetric payoff effects while preserving modularity \citep{albert2025postestimation}.  
CAM-DF extends the scope of decision-focused learning to the realm of cost-aware stopping over ranked tool prefixes by deriving a regret target from the payoff difference between the current prefix and its best continuation and establishing its Bayes alignment with the offline stopping target.

\paragraph{Costly search and stopping.}
Classical costly-search models compare the expected benefit of an additional observation with its acquisition cost. Weitzman's Pandora's-box model uses reservation values to determine a fixed inspection order and an outcome-adaptive stopping rule \citep{weitzman1979optimal}. Subsequent work considers constrained inspection orders, utilities over multiple discovered values, and contextual learning of search policies \citep{beyhaghi2023pandora,boodaghians2023pandora,olszewski2015general,atsidakou2024contextual}. Recent LLM applications use sequential stopping to control repeated generation and API cascades after observing intermediate outputs \citep{kalayci2025optimalstopping,belloni2026online}. CAM-DF introduces a pre-execution, set-level formulation of costly stopping for complementary tool access, in which the selected prefix is fixed before any tool output is observed.

\section{Problem Formulation}
\label{sec:problem}

\paragraph{Setting.}
We consider a black-box acquisition layer around a tool-using service agent within a single agent turn, using only public signals and requiring no access to the agent's internals. The layer receives a candidate set $\mathcal{T}=\{1,\dots,m\}$, an upstream ranking over those candidates, and a cost $c_j>0$ for calling each tool $j$; costs may be homogeneous ($c_j\equiv 1$) or heterogeneous. It walks the ranking, chooses a prefix, and hands that prefix to the agent to execute. The prefix is fixed before the first tool runs, so the decision rests only on the ranking, the costs, and the layer's own acquisition progress. Such a ranking is readily available in modern harnesses and may come from any upstream component that orders tools: a retriever, router, scoring prompt, workflow rule, or offline shadow-labeling pass. Each task $x$ has a minimal required-tool set $G_x\subseteq\mathcal{T}$. An acquired set $A$ is sufficient if and only if $G_x\subseteq A$. Tools in $A\setminus G_x$ do not change task sufficiency but incur additional acquisition cost. $G_x$ is a latent property of the task and is unavailable to the acquisition layer at decision time. For logged tasks, it is available through benchmark
annotations and used only offline to construct training labels and evaluation metrics.

In our empirical replay we populate the ranking interface with frozen LLM scoring prompts that see only the task text $x$ and tool descriptions $d_j$,
\begin{equation}
s_j(x) = h(x, d_j) \in [0,1], \qquad j=1,\dots,m.
\end{equation}
This scorer plays the role a retrieval or routing stage plays in deployed stacks \citep{qin2024toolllm,du2024anytool}.
The scores rank tools by estimated necessity but are not calibrated membership probabilities $\Pr(j\in G_x \mid x, d_j)$, and the problem is converting them into an acquisition decision.

\paragraph{Payoff.}
For an acquired set $A\subseteq\mathcal{T}$ on task $x$, the payoff is $U(A,x)=v(A,G_x)-\lambda\sum_{j\in A}c_j$, where $v(A,G_x)\in[0,1]$ is the task value assigned to $A$, and $\lambda>0$ is the cost pressure, measured as the price of one unit of tool cost in units of task value. Under exact sufficiency, $v(A,G_x)=\mathbf{1}\{G_x\subseteq A\}$, yielding
\begin{equation}
\Uex(A,x)
=
\mathbf{1}\{G_x\subseteq A\}
-
\lambda\sum_{j\in A}c_j.
\label{eq:uexact}
\end{equation}
%The all-or-nothing value represents settings in which a single missing required tool invalidates the downstream action. 
The indicator equals one when $A$ contains all required tools and zero otherwise. Tools acquired beyond $G_x$ add cost but no task value, whereas omitting any required tool makes $A$ insufficient. As a robustness specification, we also use partial coverage, $v(A,G_x)=|A\cap G_x|/|G_x|$, yielding
$\Upar(A,x)=|A\cap G_x|/|G_x|-\lambda\sum_{j\in A}c_j$, which assigns partial credit to acquired required tools and represents tasks whose value accumulates across multiple pieces of information (Table~\ref{tab:partial}). Both value functions are nondecreasing in $A$; acquisition costs enter separately through the penalty term.

\paragraph{Stopping along the score ranking.}
Order the tools by score, $s_{(1)}(x)\ge \cdots \ge s_{(m)}(x)$, and let $A_t=\{(1),\dots,(t)\}$ denote the prefix of the first $t$ tools.
The policy chooses a stopping depth $t\in\{0,\dots,m\}$: at each state $S_t$ (the situation after acquiring the first $t$ tools) it either stops with $A_t$ or acquires the next-ranked tool.
Restricting to score-ordered prefixes keeps the upstream interface unchanged and reduces a $2^m$ subset choice to $m{+}1$ nested decisions.

\paragraph{Observable state and model input.}
The policy input $X_t$ at state $S_t$ contains variables observable at decision time: the scores and costs of selected and remaining tools, score gaps, acquisition progress, and $\lambda$. Because $G_x$ is latent at decision time, variables derived from $G_x$ are unavailable as inputs: the required set itself, $|G_x|$, the rank needed to cover $G_x$, or missing-required indicators. We use the annotated $G_x$ only offline, to construct training labels, regret weights, and evaluation metrics.

\paragraph{Stopping target.}
At state $S_t$, the policy either stops with the current prefix $A_t$ or continues. Let the stop payoff be $\Qstop(S_t) = U(A_t,x)$ and the best-continuation payoff be $\Qcont(S_t) = \max_{k>t} U(A_k,x)$, and define the decision gap
\begin{equation}
\Delta(S_t) = \Qstop(S_t) - \Qcont(S_t).
\label{eq:delta}
\end{equation}
Stopping is optimal exactly when continuing cannot do better, i.e., when $\Delta(S_t)\ge 0$. This defines the target stop/continue label $y_t^{*}=\mathbf{1}\{\Delta(S_t)\ge 0\}$, and $|\Delta(S_t)|$ is the payoff lost by choosing the wrong action at $S_t$. For a fixed task and ranking, $y_t^*=1$ exactly when the current prefix is optimal among the current and all later prefixes. The best attainable payoff is $\max_k U(A_k,x)$, and $\Delta(S_t)$ measures how far the current prefix falls short of the best later one. Both $\Delta(S_t)$ and $y_t^*$ depend on $G_x$, so they are computable on logged tasks but not at the decision time (Lemma \ref{lem:frontier} in the appendix gives the deterministic characterization of the optimal depth). 

\section{Theory}
\label{sec:theory}

One alignment theorem and three structural propositions (proofs in Appendix~\ref{app:proofs}) pin down what a learned stopping policy must represent and which features carry the decision.

\subsection{From Local Errors to Prefix Regret}
For a fixed task and ranking, let $F_t=\max_{k\ge t}U(A_k,x)$ denote the highest payoff attainable from depth $t$ onward. A stopping classifier $g(X_t)\in\{-1,+1\}$ assigns $+1$ to stopping and $-1$ to continuing. It induces the stopping depth $\tau
=
\min\left(
\{t<m:g(X_t)=+1\}\cup\{m\}
\right).
$
For each nonterminal state, let $Y_t=+1$ when $\Delta(S_t)\ge0$ and $Y_t=-1$ otherwise. When $\Delta(S_t)=0$, stopping and continuing attain the same best payoff, and we assign the tie to stopping. The realized prefix regret is
$
\operatorname{Regret}(g)
=
F_0-U(A_\tau,x).
$

\begin{theorem}[Pathwise Prefix-Regret Decomposition]
\label{thm:payoff}
For every task $x$ and stopping classifier $g$,
\begin{equation}
\operatorname{Regret}(g)
=
\sum_{t=0}^{\min\{\tau,m-1\}}
|\Delta(S_t)|
\mathbf{1}\{g(X_t)\neq Y_t\}.
\label{eq:pathwise}
\end{equation}
\end{theorem}

Theorem~\ref{thm:payoff} shows that realized regret is a sum of weighted mistakes, which motivates training against the regret-weighted classification risk defined below:
\begin{equation}
\mathcal{R}(g) = \mathbb{E}\bigl[\, |\Delta(S_t)| \cdot \mathbf{1}\{g(X_t)\neq Y_t\} \,\bigr].
\label{eq:risk}
\end{equation}

% A stopping policy is a classifier $g(X_t)\in\{-1,+1\}$ ($+1=$ stop) that induces the stopping time $\tau=\min\{t:g(X_t)=+1\}$ (with $\tau=m$ if $g$ never stops). Let $Y_t=+1$ if $\Delta(S_t)\ge 0$ and $Y_t=-1$ otherwise, the payoff-optimal stop/continue action at $S_t$. Define $F_0=\max_k U(A_k,x)$ as the prefix-oracle payoff, and the realized regret of policy $g$ is $\operatorname{Regret}(g) = F_0-U(A_{\tau},x)$.
% \begin{theorem}[Pathwise Prefix-Regret Decomposition]
% \label{thm:payoff}
% For every task $x$ and every stopping policy $g$, 
% \begin{equation}
% \operatorname{Regret}(g) = \sum_{t=0}^{\tau} |\Delta(S_t)| \cdot \mathbf{1}\{g(X_t)\neq Y_t\} .
% \label{eq:pathwise}
% \end{equation}
% \end{theorem}
% The regret of $g$ equals the sum of decision gaps $|\Delta(S_t)|$ at the states where $g$ disagrees with the optimal action, and vanishes iff $g$ agrees with $Y_t$ at all visited states. 

\subsection{Bayes Alignment of the Regret-Weighted Target}

\begin{theorem}[Bayes alignment for clairvoyant labels]
\label{thm:bayes}
Any Bayes minimizer of $\mathcal{R}(g)$ satisfies, almost surely on the event
$\{\mathbb{E}[\Delta(S_t)\mid X_t]\neq 0\}$,
\[
g^{*}(X_t) =
\operatorname{sign}\bigl(\mathbb{E}[\Delta(S_t)\mid X_t]\bigr)
\]
\end{theorem}

Theorem~\ref{thm:bayes} shows that weighting each error by the payoff at stake $|\Delta(S_t)|$ turns classification into payoff optimization. This is the Bayes rule of cost-sensitive learning \citep{elkan2001cost}, here instantiated for decision-focused stopping \citep{elmachtoub2022spo}. It targets a different object than our predict-then-threshold baseline (Section~\ref{sec:method}), which estimates current sufficiency $\Pr(G_x\subseteq A_t\mid X_t)$ and stops once predicted sufficiency clears a threshold. A perfectly calibrated sufficiency predictor can still stop at the wrong depth when cost is involved. In the training, we replace 0--1 loss with logistic loss while keeping the $|\Delta(S_t)|$ weights. The bounded weights act as a change of measure, so logistic calibration maintains the target decision boundary \citep{bartlett2006convexity,scott2012calibrated,elkan2001cost}. Lemma~\ref{lem:logistic} in the appendix  states the calibrated-surrogate result. We also note that Theorem~\ref{thm:bayes} is an alignment result for the training target: it uses the clairvoyant gap $\Delta(S_t)$ and scores all prefix states, whereas the deployed policy sees only $X_t$ and stops at the first predicted stop. We let the experiments measure the performance in the policy-level.

\subsection{Marginal Value and the Limits of Sufficiency Prediction}

Theorem~\ref{thm:bayes} shows the optimal policy tracks $\operatorname{sign}\bigl(\mathbb{E}[\Delta(S_t)\mid X_t]\bigr)$, while the following three results answer which observable features carry that sign. Proposition~\ref{prop:marginal} reduces the stop/continue call to a marginal value-per-cost test, identifying the quantities a policy must estimate; Proposition~\ref{prop:index} solves the decision in closed form under an additive-value benchmark; and Proposition~\ref{prop:fail} shows that this decision cannot be recovered from relevance scores alone once costs are heterogeneous. Together they imply what CAM-DF's features must encode and why score-only rules are structurally insufficient.

\begin{proposition}[One-step marginal rule]
\label{prop:marginal}
\normalfont
For a nonterminal state $S_t$, let $j$ be the tool at rank $t+1$. Conditional on the observable features $X_t$, define
%Conditional on the deployment-observable features $X_t$, let $j$ denote the tool at rank $t+1$ and define
\[
V_t=\mathbb{E}[v(A_t,G_x)\mid X_t],
\quad
V_{t+1}=\mathbb{E}[v(A_t\cup\{j\},G_x)\mid X_t].
\]
When comparing stopping at $A_t$ with acquiring $j$ and then stopping, continuing is preferred if and only if $\frac{V_{t+1}-V_t}{c_j}\ge\lambda$.
\end{proposition}

This proposition is a one-step comparison of the optimal prefix. Under the exact-sufficiency payoff $\Uex$,
\[
V_{t+1}-V_t
=
\Pr(G_x\setminus A_t=\{j\}\mid X_t),
\]
the probability that the next tool completes the required set. Under this payoff, current-prefix sufficiency $V_t$, even if known exactly, does not by itself determine the preferred action: holding $V_t$ fixed, differences in the completion probability or the next-tool cost can reverse the decision. Under the partial-coverage payoff $\Upar$, $V_{t+1}-V_t$ is the expected normalized coverage contributed by the next tool. In both cases the decision depends on the selected prefix, the next tool, and the remaining candidates.

% \begin{proposition}[Additive index]
% \label{prop:index}
% \normalfont
% Suppose value is additive, $\mathbb{E}[v(A,G_x)\mid x] = \sum_{j\in A}p_j(x)$ with $p_j(x) = \Pr(j\in G_x\mid x)$, or the analogous partial-coverage normalization. Over \emph{unconstrained subsets} (not score-ordered prefixes), the optimal acquisition rule is separable:
% \[
% j\in A^{*}(x)
% \quad\Longleftrightarrow\quad
% \frac{p_j(x)}{c_j}\ge \lambda .
% \]
% \end{proposition}

% Under homogeneous costs the rule collapses to a posterior threshold $p_j(x)\ge\lambda$, which explains why tuned score thresholds are strong baselines in homogeneous-cost settings. \edit{The additive model is a highly stylized assumption. The above threshold policy only works if additive index is in force, which is not applicable in general.}

\begin{proposition}[Additive acquisition structure]
\label{prop:index}
\normalfont
Suppose the expected non-cost value is additive,
$
\mathbb{E}[v(A,G_x)\mid x] \allowbreak
=
\sum_{j\in A}p_j(x),
$
where $p_j(x)$ is the standalone value of tool $j$. Under unconstrained subset acquisition, an optimal set keeps every tool whose value-per-cost clears the pressure,
\[
A^{*}(x)
=
\left\{
j:
\frac{p_j(x)}{c_j}\ge\lambda
\right\}.
\]

Under the fixed score ordering, the optimal prefix depth is 
\[
t^{*}(x)
\in
\arg\max_{0\le t\le m}
\sum_{r=1}^{t}
\left(
p_{(r)}(x)-\lambda c_{(r)}
\right).
\]
Consequently, at a nonterminal prefix $A_t$, continuing to some later prefix is strictly preferred if and only if
\[
\max_{k>t}
\sum_{r=t+1}^{k}
\left(
p_{(r)}(x)-\lambda c_{(r)}
\right)
>0.
\]
\end{proposition}

With homogeneous costs and an ordering aligned with $p_j(x)$, the fixed-prefix rule reduces to a value threshold, explaining why tuned score thresholds are strong in homogeneous-cost settings. Under heterogeneous costs, however, value-per-cost ratios need not follow the score ordering, and the optimal stopping depth depends on the cumulative surplus of later candidates. This structure motivates CAM-DF's remaining tool score and cost features. Additivity is a tractable case with closed-form solution; the exact-sufficiency payoff instead couples tools through set completion, motivating learning rather than a fixed index policy.

\begin{proposition}[Failure of score-monotone acquisition]
\label{prop:fail}
\normalfont
Under the additive model and the unconstrained-subset decision in Proposition~\ref{prop:index}, suppose two tools satisfy
\[
p_i(x)>p_j(x)
\qquad\text{but}\qquad
\frac{p_i(x)}{c_i}<\frac{p_j(x)}{c_j}.
\]
For any $\lambda\in \left( \frac{p_i(x)}{c_i}, \frac{p_j(x)}{c_j} \right)$, the optimal cost-aware subset includes tool $j$ and excludes tool $i$. Any rule monotone in value scores alone must include $i$ whenever it includes $j$ and therefore cannot reproduce this decision. Thus, under heterogeneous costs, score-only acquisition can be inconsistent with the cost-aware optimum.
\end{proposition}

Together, Propositions~\ref{prop:index} and~\ref{prop:fail} show that acquisition decisions depend on marginal value relative to cost and, under a fixed ranking, on the cumulative net value of the remaining candidates. This structure motivates CAM-DF's joint use of score, cost, and progress features.

% Propositions~\ref{prop:marginal}--\ref{prop:fail} are deliberately simple relatives of reservation-price indices in sequential search \citep{weitzman1979optimal} and the Pandora's-box family \citep{beyhaghi2023pandora,atsidakou2024contextual}; our setting has set-valued success, noisy LLM scores, and $\lambda$-weighted heterogeneous costs.
% Proposition~\ref{prop:fail} \edit{proposes a scenario where score-only \emph{orderings} fails to find the optimal tool acquisition.} CAM-DF learns only stopping depth; Sec.~\ref{sec:exp-order} shows re-ordering by $s_j/c_j$ leaves payoff statistically unchanged at $\lambda=0.12$.

\subsection{Theory-Suggested Hypotheses}

The structure above suggests when decision-focused stopping should pay. If scores perfectly separated the required from the non-required tools, a simple threshold would suffice; if they carried no signal, no stopping rule could help. In between, the gap between a payoff-aligned rule and a predict-then-threshold or score-only rule should widen with: (P1) score ambiguity, i.e. lower scorer AUC;
(P2) cost dispersion, since varying costs make value-per-cost order diverge from score order (Propositions~\ref{prop:index} and \ref{prop:fail});
(P3) cost pressure $\lambda$, since larger $\lambda$ raises the regret of each unnecessary acquisition.
Sections~\ref{sec:exp-main}--\ref{sec:exp-robust} test P1--P3 directly.

\section{Method: CAM-DF and CAM-DF-lite}%
\label{sec:method}%

The theory motivates two components of CAM-DF. The sign and magnitude of the decision gap define the stop-or-continue label and its regret weight, while the marginal analysis motivates features that capture additional value relative to acquisition cost. CAM-DF combines these components in a supervised stopping classifier.

\paragraph{CAM-DF.}

CAM-DF operationalizes Theorem~\ref{thm:bayes} as an $\ell_2$-regularized, regret-weighted logistic classifier. For each nonterminal prefix $S_t$, $t=0,\ldots,m-1$, the stop label is $y_t^*=\mathbf{1}\{\Delta(S_t)\ge0\}$ and the sample weight is $w_t=|\Delta(S_t)|+\epsilon$, where $\epsilon=10^{-4}$ keeps zero-gap examples in training. Errors with larger payoff consequences therefore receive greater emphasis. Pooling the nonterminal prefixes from all training tasks yields $N$
training examples, each corresponding to one task at a particular
candidate depth. CAM-DF models
$p_\theta^{\mathrm{stop}}(X_n)=\sigma(\theta_0+\theta^\top\phi(X_n))$ and minimizes
\begin{equation}
\widehat{\mathcal{L}}(\theta)
=
\frac{1}{N}\sum_{n=1}^{N}
\bigl(|\Delta_n|+\epsilon\bigr)
\ell_{\mathrm{log}}
\left(y_n^*,p_\theta^{\mathrm{stop}}(X_n)\right)
+
\gamma\|\theta\|_2^2,
\label{eq:camdf-loss}
\end{equation}
where $\Delta_n$ and $y_n^*$ denote the payoff gap and label of example $n$, $\sigma$ is the sigmoid function, $\ell_{\mathrm{log}}$ is binary logistic loss, and $\gamma$ controls regularization. The feature map uses only scores, costs, prefix progress, and next-tool information available at deployment. The required-set annotation $G_x$ enters training through the labels and weights, but not through the feature map. Further implementation details are summarized in Appendices~\ref{app:setup}--\ref{app:algo}.

At deployment, CAM-DF scans candidate depths in increasing order and stops at the first state satisfying $p_\theta^{\mathrm{stop}}(X_t)\ge0.5$. If no earlier state satisfies this condition, it selects the full ranked list. The scan occurs before execution and reveals no tool output. The downstream agent receives only the selected prefix.

\paragraph{CAM-DF-lite.}
CAM-DF-lite is a lower-dimensional, interpretable variant of CAM-DF. Different from CAM-DF's full feature map, it uses ten theory-motivated marginal score-cost and prefix-progress features. It predicts $p_\theta^{\mathrm{cont}}(X_t)=\Pr_\theta(\Delta(S_t)<0\mid X_t)$ and stops when this probability falls below a validation-tuned threshold, whereas CAM-DF uses a fixed $0.5$ stop gate. The decision target, loss, and prefix-scanning procedure are unchanged. Positive coefficients therefore favor further acquisition (Table~\ref{tab:coef}).

\section{Experiments}%
\label{sec:experiments}%

\paragraph{Datasets.}
We use Retail as the primary benchmark and four additional domains to assess cross-domain generality (Table~\ref{tab:datasets}) \citep{yao2024taubench,mcpatlas2026,styles2024workbench,barres2025tau2}.
\begin{table}[!htbp]
\centering
\small
\setlength{\tabcolsep}{3pt}
\begin{tabular}{@{}p{0.29\columnwidth}p{0.52\columnwidth}r@{}}
\toprule
Dataset & Evaluation role & Tasks \\
\midrule
$\tau$-bench Retail & Main testbed with audited required-tool annotations. & 67 \\
MCP-Atlas & Larger candidate catalogs with trajectory-derived labels. & 495 \\
WorkBench & Larger-scale outcome-centric evaluation. & 639 \\
$\tau^2$-bench Airline & Small-domain, low-power evaluation. & 28 \\
$\tau^2$-bench Telecom & Second conversational-agent evaluation. & 114 \\
\bottomrule
\end{tabular}
\caption{Evaluation domains (1{,}343 tasks in total). Label provenance and per-domain protocols are in Appendices~\ref{app:setup} and~\ref{app:crossdomain}.}%
\label{tab:datasets}%
\end{table}

\paragraph{Baselines.}
We compare against six baselines, each isolating a different acquisition signal. All tunable parameters are selected on validation payoff only.
\begin{enumerate}
\setlength{\itemsep}{0pt}
\setlength{\parsep}{0pt}
\setlength{\parskip}{0pt}
\item \textbf{Tuned fixed-$k$:} acquires the same validation-tuned number $k$ of top-ranked tools for every task.
\item \textbf{Score threshold:} acquires all tools with $s_j\ge\tau$.
\item \textbf{Score-per-cost threshold:} independently acquires each tool satisfying $s_j/c_j\ge\tau$, providing a simple cost-aware heuristic motivated by Proposition~\ref{prop:index}.
\item \textbf{Retail rule:} selects tools using a hand-written mapping from request keywords and supplements this set with candidates passing a score threshold.
\item \textbf{Predict-then-threshold:} predicts whether the current prefix is sufficient and stops when exceeding a validation-tuned threshold \citep{jeong2024adaptiverag}.
\item \textbf{Aggregate-only DF:} uses the CAM-DF objective with only aggregate summaries of the selected and remaining candidates, excluding next-tool and tool-identity features.
\end{enumerate}
For reference, we also report a nondeployable \emph{prefix oracle} that uses the true required set $G_x$ to select the payoff-maximizing ranked prefix for each task, providing an upper bound for prefix-stopping policies.

\paragraph{Experimental setup.}
The evaluation uses four shared settings:
\begin{enumerate}
\setlength{\itemsep}{0pt}
\setlength{\parsep}{0pt}
\setlength{\parskip}{0pt}
\item \textbf{Rankings:} qwen-plus provides the primary rankings, which remain fixed during policy training and evaluation. Four additional frozen scorer configurations assess sensitivity to ranking quality.
\item \textbf{Cost parameters:} the main Retail analysis uses a $3\times3$ grid defined by cost pressure $\lambda\in\{0.05,0.12,0.20\}$ and cost dispersion $d\in\{0,1.0,1.5\}$, where $d$ controls the spread of costs around their mean and $d=0$ is uniform.
\item \textbf{Tuning:} methods are selected by validation payoff over 30 random 55/20/25\% train/validation/test splits (Atlas: 20 seeds; airline: 50; Appendix~\ref{app:setup}).
\item \textbf{Evaluation:} we report payoff, sufficiency, acquisition cost, and paired task-bootstrap 95\% confidence intervals using 2{,}000 resamples. All results except the live evaluation are offline replays.
\end{enumerate}
Additional protocols and implementation details are provided in Appendices~\ref{app:setup} and~\ref{app:crossdomain}.

\subsection{Main Results}%
\label{sec:exp-main}%
\label{sec:exp-cross-domain}%

\begin{table*}[t]
\centering
\small
\setlength{\tabcolsep}{3pt}
\renewcommand{\arraystretch}{1.03}
\begin{tabular}{@{}lccc@{\hspace{6pt}}cccc@{}}
\toprule
& \multicolumn{3}{c}{$\tau$-bench Retail ($n{=}67$), exact payoff}
& \multicolumn{4}{c}{Additional benchmarks, heterogeneous costs ($d{=}1.0$)} \\
\cmidrule(lr){2-4}\cmidrule(lr){5-8}
Policy & $d{=}0$ & $d{=}1.0$ & $d{=}1.5$
& MCP-Atlas & WorkBench & Telecom & Airline \\
& & & & {\small $n{=}495$} & {\small $n{=}639$} & {\small $n{=}114$} & {\small $n{=}28$} \\
\midrule
Tuned fixed-$k$        & $0.277^{\dagger}$ & $0.219^{\dagger}$ & $0.196^{\dagger}$ & $0.200^{\dagger}$ & $0.593^{\dagger}$ & $-0.006$ & $0.484$ \\
Score threshold        & $0.376$ & $0.302$ & $0.267$ & $0.201^{\dagger}$ & $0.626^{\dagger}$ & $-0.157^{\dagger}$ & $0.488$ \\
Score-per-cost thr.\   & $0.376$ & $0.293$ & $0.273$ & $0.216^{\dagger}$ & $0.644^{\dagger}$ & $-0.066^{\dagger}$ & $\mathbf{0.507}$ \\
Retail rule            & $0.350$ & $0.298$ & $0.272$ & --- & --- & --- & --- \\
Predict-then-threshold & $0.352^{\dagger}$ & $0.281^{\dagger}$ & $0.244^{\dagger}$ & $0.195^{\dagger}$ & $0.617^{\dagger}$ & $-0.173^{\dagger}$ & $0.486$ \\
Aggregate-only DF      & $0.369$ & $0.275^{\dagger}$ & $0.246^{\dagger}$ & $0.204^{\dagger}$ & $0.624^{\dagger}$ & $-0.017$ & $0.493$ \\
CAM-DF-lite            & $0.383$ & $0.305$ & $0.276$ & $\mathbf{0.226}$ & --- & --- & --- \\
CAM-DF (ours)          & $\mathbf{0.400}$ & $\mathbf{0.331}$ & $\mathbf{0.298}$ & --- & $\mathbf{0.660}$ & $\mathbf{-0.001}$ & $0.483$ \\
\midrule
Prefix oracle          & $0.513$ & $0.450$ & $0.419$ & $0.337$ & $0.772$ & $0.154$ & $0.727$ \\
\bottomrule
\end{tabular}
\caption{The main benchmark is retail at $\lambda{=}0.12$ ; external domains use heterogeneous costs ($d{=}1.0$). Atlas reports 20-seed partial payoff with CAM-DF-lite (Aggregate-only DF is its cost-blind control); all other columns use exact payoff with full CAM-DF. Bold: best deployable method; $^{\dagger}$: paired CAM-policy gain with a 95\% CI above zero.}%
\label{tab:main}%
\end{table*}

\begin{figure}[tb]
\centering
\includegraphics[width=\columnwidth]{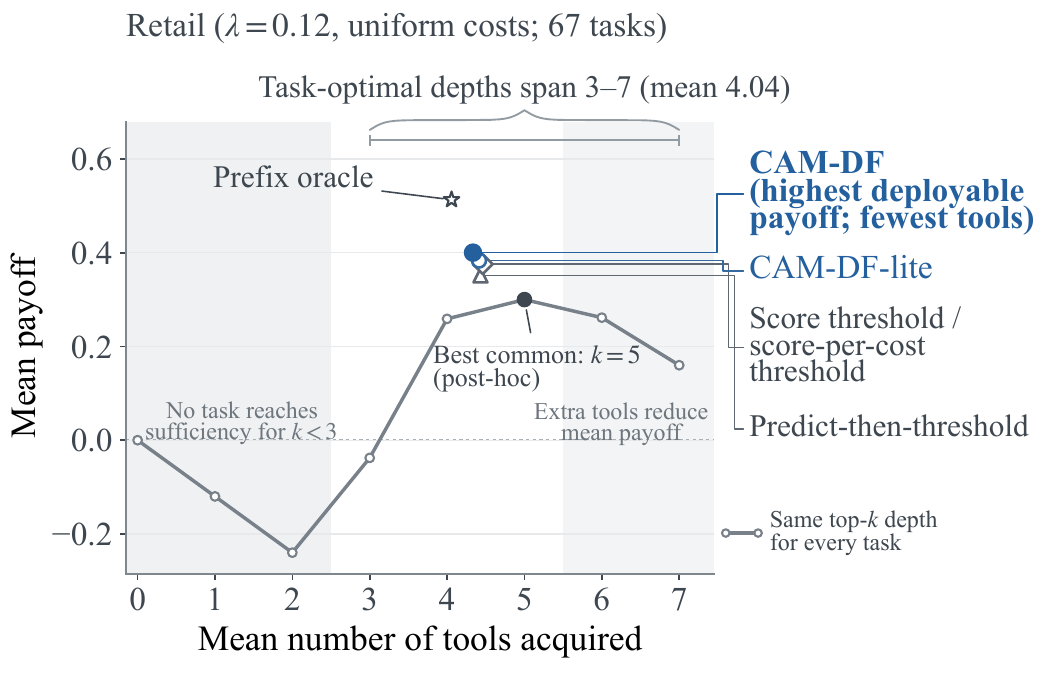}
\caption{Payoff versus tool acquisition on Retail
($\lambda{=}0.12$, $d{=}0$). The gray curve uses a common
top-$k$ depth for all tasks, while markers show task-specific
policies at their mean tool count and payoff. Shading indicates
under- and over-acquisition.}
\label{fig:payoffdepth}
\end{figure}

\paragraph{Performance across domains.}
Table~\ref{tab:main} reports the primary comparison at $\lambda=0.12$. On Retail, CAM-DF achieves the highest mean payoff at every tested cost dispersion and significantly outperforms Tuned fixed-$k$ and Predict-then-threshold throughout, setting a new state of the art among deployable tool-acquisition methods on this benchmark. The score-based thresholds remain competitive at this moderate cost pressure but do not match CAM-DF in mean payoff. CAM-DF-lite ranks second across all three Retail settings and first on MCP-Atlas, while CAM-DF ranks first on WorkBench and Telecom under heterogeneous costs. Airline remains inconclusive at $n=28$, with the Score-per-cost threshold highest in mean. Figure~\ref{fig:payoffdepth} clarifies why task-specific stopping improves on a common acquisition depth. Under uniform costs, the post-hoc best common depth is $k=5$, whereas task-optimal depths range from three to seven. CAM-DF achieves a mean payoff of $0.400$ with $4.34$ tools on average, exceeding the common-depth maximum of $0.300$, while the prefix-oracle payoff of $0.513$ indicates remaining headroom.

\begin{figure}[!htbp]
\centering
\includegraphics[width=\columnwidth]{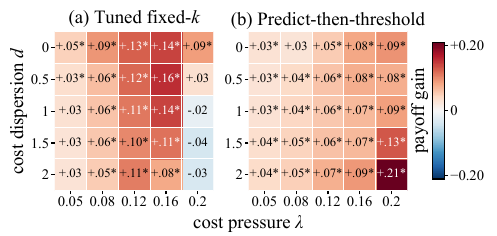}
\caption{Retail payoff gains of CAM-DF over (a) Tuned fixed-$k$ and (b) Predict-then-threshold across cost pressure $\lambda$ and dispersion $d$. Red favors CAM-DF and blue the baseline; asterisks denote paired-bootstrap 95\% CIs above zero.}
\label{fig:regimemap}
\end{figure}

\paragraph{Cost pressure and heterogeneity.}
For comparative statics, we evaluate a denser $5\times5$ grid. CAM-DF outperforms Predict-then-threshold in mean in all 25 regimes, with paired-bootstrap 95\% CIs above zero in 24. Gains increase consistently with $\lambda$ and become largest at high $d$, supporting P3 and providing directional support for P2; the peak is $+0.21$ at $\lambda{=}0.20$ and $d{=}2$. Gains over Tuned fixed-$k$ narrow or reverse at the highest cost pressure, where a conservative common depth remains competitive (Figure~\ref{fig:regimemap}; Table~\ref{tab:griddense}).

\subsection{Ablation Study}%
\label{sec:exp-ablation}%

\paragraph{Component and objective ablations.}
The ablations separate the roles of state representation, the $\Delta$-based objective, regret weighting, and model complexity. Restricting CAM-DF to aggregate prefix features lowers payoff by $0.034$--$0.051$, with CIs excluding zero under heterogeneous costs (Table~\ref{tab:ablation}). Directly regressing $\Delta$ performs comparably to regret-weighted classification, with paired differences ranging from $-0.006$ to $+0.025$ and all CIs covering zero, while remaining above Predict-then-threshold in mean (Appendix~\ref{app:robust}). This suggests that the main signal comes from targeting the stop-versus-continue payoff gap rather than from choosing classification over regression. Removing regret weights lowers payoff by $0.034$--$0.041$ at $\lambda{=}0.12$, although the CIs cover zero, and by $0.072$--$0.079$ at $\lambda{=}0.05$, where the CIs exclude zero. CAM-DF-lite remains within $0.022$--$0.027$ of the full model, with all CIs covering zero. The no-identity variant also retains mean gains of $+0.035/+0.045$ over Predict-then-threshold at $d{=}1.0/1.5$. These results indicate that the gains are associated with $\Delta$-targeted stopping and richer state information, without depending solely on model complexity or tool identity.

\paragraph{Value of learning across cost regimes.}
We next examine when learned stopping improves on simple acquisition rules. At $\lambda{=}0.12$, CAM-DF significantly outperforms the direct plug-in $s_j\ge\lambda c_j$ by $0.059$--$0.071$, while the fixed score rule $s_j\ge0.5$ remains statistically competitive. At $\lambda{=}0.20$ and $d{=}1.5$, however, all four training-free rules lose significantly by $0.100$--$0.212$ (Table~\ref{tab:trainfree}). This boundary also appears on WorkBench and Telecom: the tuned Score-per-cost threshold is competitive under uniform costs, but CAM-DF leads under heterogeneous costs by $0.008$--$0.014$ and $0.067$--$0.085$, respectively (Tables~\ref{tab:workbench} and~\ref{tab:tau2}). Additional controls show that CAM-DF's advantage over Predict-then-threshold is robust to cost-vector permutations and symmetric tuning protocols (Appendix~\ref{app:robust}).

\subsection{Robustness and End-to-End Evaluation}%
\label{sec:exp-robust}

\paragraph{Ranking-source robustness.}
To test whether CAM-DF's advantage depends on the qwen-plus ranking used in the main analysis, we rerun the full training and tuning pipeline under five frozen scorer configurations with required-tool AUCs ranging from $0.835$ to $0.953$. At $\lambda{=}0.12$, each scorer is evaluated under uniform and heterogeneous costs, yielding ten comparisons. CAM-DF exceeds the feature-matched, cost-inclusive Predict-then-threshold baseline in mean in all ten comparisons, with 95\% CIs excluding zero in nine (Table~\ref{tab:costaware}). Gains are largest for the lowest-AUC scorer but do not vary monotonically across all five configurations, providing directional support for P1. Against the original score-only baseline, all ten gains have CIs excluding zero (Table~\ref{tab:cross}).

% \paragraph{Prediction and label quality.}
% We test whether CAM-DF's gain can be explained by errors in sufficiency prediction and how it depends on required-set labels. A cross-fitted residual correction \citep{chernozhukov2018dml} does not materially improve the sufficiency baseline, suggesting that correctable miscalibration alone does not explain the gain. In a separate sensitivity test, the learned advantage disappears when $10\%$ of required-set memberships are corrupted. Reliable required-set labels therefore remain an important training requirement (Appendix~\ref{app:robust}; Table~\ref{tab:noise}).

\paragraph{End-to-end validation.}
We test whether required-set sufficiency predicts realized task success and whether the learned policies reduce tool exposure in actual agent execution. On $\tau$-bench Retail, providing the annotated required set yields $0.67$ success, whereas removing one required tool lowers success to $0.24$ ($p{<}0.001$), supporting the offline sufficiency proxy. CAM-DF and CAM-DF-lite achieve success rates of $0.67$ and $0.64$ while exposing only $4.4$ and $4.3$ of seven read tools, a $37\%$--$39\%$ reduction from full access, whose success rate is $0.58$. Neither policy differs detectably from full access (McNemar $p{=}0.21/0.42$), although these tests do not establish non-inferiority. An independent user simulator reproduces the pattern (Tables~\ref{tab:e2e}--\ref{tab:e2eusim}).

\section{Conclusion} \label{sec:conclusion}%
This paper formulates pre-execution tool acquisition as a cost-aware stopping problem over ranked candidates. The CAM-DF family learns acquisition depth from stop-versus-continue payoff gaps through a regret-weighted objective with Bayes-alignment guarantees, and the marginal analysis characterizes when score-only acquisition is suboptimal. Experiments across five domains document payoff improvements across multiple ranking sources and cost regimes. The live Retail evaluation further demonstrates lower tool exposure with comparable observed task success. Together, these findings support cost-aware stopping as a modular control layer that converts an existing ranking into a decision without modifying the upstream ranker or underlying LLM. While CAM-DF provides a cost-aware framework for one-turn pre-execution acquisition, the current formulation does not update acquisition decisions after observing tool outputs. Extending the framework to output-adaptive multi-round settings always deserves attention.

\bibliography{references}

\newpage
\appendix
\twocolumn[{%
 \centering
 \LARGE \textbf{Supplementary Material:\\ Scores Are Not Decisions: Cost-Aware Stopping for Tool Acquisition in LLM Agents}\\[2em]
}]

\renewcommand{\thetable}{A\arabic{table}}
\renewcommand{\thefigure}{A\arabic{figure}}
\renewcommand{\thealgorithm}{A\arabic{algorithm}}
\setcounter{table}{0}
\setcounter{figure}{0}

\section{Proofs}%
\label{app:proofs}%

\subsection{Proof of Theorem~\ref{thm:payoff}}
Write $U_t=U(S_t,x)$ and $F_t=\max_{k\ge t}U_k$. For every
$t<\tau$, the policy continues. If $Y_t=+1$, then
$F_t=U_t$ and
\[
|\Delta(S_t)|
=
U_t-F_{t+1}
=
F_t-F_{t+1}.
\]
Because continuing disagrees with the stop label, this quantity is
exactly the regret contribution at state $t$. If $Y_t=-1$, then
$F_t=F_{t+1}$, and continuing incurs no regret contribution. Thus,
for every $t<\tau$,
\[
|\Delta(S_t)|\mathbf{1}\{g(X_t)\neq Y_t\}
=
F_t-F_{t+1}.
\]
Summing over the continuation states gives
\[
\sum_{t=0}^{\tau-1}
|\Delta(S_t)|\mathbf{1}\{g(X_t)\neq Y_t\}
=
F_0-F_\tau.
\]

If $\tau<m$, the policy stops at $S_\tau$. When $Y_\tau=+1$,
we have $F_\tau=U_\tau$, so stopping incurs no regret. When
$Y_\tau=-1$, we have $F_\tau=F_{\tau+1}>U_\tau$, and
\[
|\Delta(S_\tau)|
=
F_{\tau+1}-U_\tau
=
F_\tau-U_\tau.
\]
Therefore,
\[
|\Delta(S_\tau)|
\mathbf{1}\{g(X_\tau)\neq Y_\tau\}
=
F_\tau-U_\tau.
\]
If $\tau=m$, there is no terminal decision term and
$F_m=U_m$. Combining the continuation and stopping terms yields
\[
F_0-U_\tau
=
\sum_{t=0}^{\min\{\tau,m-1\}}
|\Delta(S_t)|
\mathbf{1}\{g(X_t)\neq Y_t\},
\]
which is the claimed identity.
\hfill$\square$

\subsection{Proof of Theorem~\ref{thm:bayes}}

Fix a feature value $X_t = x$.
If the classifier stops, the conditional risk is the expected weighted error incurred on states where continuing is better:
\begin{equation*}
R_{\mathrm{stop}}(x) = \mathbb{E}\bigl[\,|\Delta(S_t)|\,\mathbf{1}\{\Delta(S_t)<0\} \mid X_t=x\,\bigr].
\end{equation*}
If it continues,
\begin{equation*}
R_{\mathrm{cont}}(x) = \mathbb{E}\bigl[\,|\Delta(S_t)|\,\mathbf{1}\{\Delta(S_t)\ge 0\} \mid X_t=x\,\bigr].
\end{equation*}
The Bayes rule stops iff $R_{\mathrm{stop}}(x) \le R_{\mathrm{cont}}(x)$, i.e.
\begin{multline*}
\mathbb{E}\bigl[-\Delta(S_t)\,\mathbf{1}\{\Delta(S_t)<0\} \mid X_t=x\bigr] \\
\le \mathbb{E}\bigl[\Delta(S_t)\,\mathbf{1}\{\Delta(S_t)\ge 0\} \mid X_t=x\bigr].
\end{multline*}
Rearranging, the two indicator terms combine into the full conditional expectation, so stopping is optimal iff $\mathbb{E}[\Delta(S_t)\mid X_t=x]\ge 0$.
Hence any Bayes minimizer of $\mathcal{R}(g)$ equals $\operatorname{sign}(\mathbb{E}[\Delta(S_t)\mid X_t])$. $\square$

\subsection{Deterministic Prefix Stopping Frontier}

\begin{lemma}[Finite-prefix stopping frontier]%
\label{lem:frontier}%
Fix a task $x$, a candidate order, and a payoff $U(A_k,x)$ over prefixes $A_0,\ldots,A_m$.
Let $F_t=\max_{k\ge t}U(A_k,x)$ be the best payoff attainable from state $t$ onward, with $F_m=U(A_m,x)$.
At a nonterminal state $t<m$, stopping is optimal among all later prefix depths iff
\[
U(A_t,x)\ge F_{t+1},
\]
equivalently iff $\Delta(S_t)=U(A_t,x)-\max_{k>t}U(A_k,x)\ge0$.
The earliest optimal stopping depth is therefore $\min\{t:U(A_t,x)=F_t\}$.
\end{lemma}

\begin{proofenv}
The feasible continuation depths from state $t$ are exactly the finite set $\{t,t+1,\ldots,m\}$.
Stopping chooses depth $t$ and obtains $U(A_t,x)$.
Continuing at least once restricts the final depth to $\{t+1,\ldots,m\}$ and obtains at best $F_{t+1}$.
Thus stopping is optimal at $t$ iff its payoff is no smaller than the best payoff from all later prefixes.
The stated equivalence follows from the definition of $\Delta(S_t)$, and the earliest optimal depth is the first index at which the running frontier $F_t$ is attained by the current prefix.
\end{proofenv}

\subsection{Weighted Logistic Surrogate}

\begin{lemma}[Logistic surrogate boundary]%
\label{lem:logistic}%
Let $Z=\Delta(S_t)$ and $W=|Z|$.
Define $Y\in\{-1,+1\}$ by $Y=+1$ iff $Z\ge0$, and assume $\mathbb{E}[W\mid X]<\infty$.
Fix $X=x$ and write $a=\mathbb{E}[W\mathbf{1}\{Y=+1\}\mid X=x]$, $b=\mathbb{E}[W\mathbf{1}\{Y=-1\}\mid X=x]$; assume $a,b>0$, so the conditional weighted logistic risk has a finite minimizer $f_x$.
If $\mathbb{E}[Z\mid X=x]\neq0$, then
\[
\operatorname{sign}(f_x)=\operatorname{sign}\{\mathbb{E}[Z\mid X=x]\}.
\]
When instead $a=0$ or $b=0$ (conditionally pure labels), no finite minimizer exists; every sequence approaching the infimum has the stated sign eventually, and the $\ell_2$ regularization used in the implementation restores a finite minimizer with that sign.
\end{lemma}

\begin{proofenv}
For fixed $x$, write
$a=\mathbb{E}[W\mathbf{1}\{Y=+1\}\mid X=x]$ and
$b=\mathbb{E}[W\mathbf{1}\{Y=-1\}\mid X=x]$.
With $\ell_y(f)=\log\{1+\exp(-y f)\}$, the conditional risk $\mathbb{E}[W\ell_Y(f)\mid X=x]$ is
$a\log(1+e^{-f})+b\log(1+e^f)$.
When $a,b>0$, its derivative is
$-a/(1+e^f)+b e^f/(1+e^f)$, so the unique minimizer satisfies $e^f=a/b$ and has sign $\operatorname{sign}(b)$.
If $a=0$ or $b=0$, the infimum is approached at the corresponding infinite sign, which is why the lemma statement assumes $a,b>0$.
Since $b=\mathbb{E}[Z\mid X=x]$, the weighted logistic population boundary matches the regret-weighted Bayes boundary.
Finite-sample regularization affects estimation, not this population target.
The $\epsilon=10^{-4}$ weight floor of Algorithm~\ref{alg:frontier} (numerical stability only) perturbs this population boundary by at most $O(\epsilon)$: the implemented weights $|Z|+\epsilon$ shift $b$ to $\mathbb{E}[Z\mid X=x]+\epsilon\,(2\Pr(Z\ge0\mid X=x)-1)$.
\end{proofenv}

\subsection{Proof of Proposition~\ref{prop:marginal}}

The two candidate actions yield expected payoffs
$V_t - \lambda\, C_t$ (stop now) and $V_{t+1} - \lambda\,(C_t + c_j)$ (acquire $j$, then stop), where $C_t$ is the cost already sunk in $A_t$.
Continuing is weakly better iff
$V_{t+1} - \lambda C_t - \lambda c_j \ge V_t - \lambda C_t$, i.e.\ $V_{t+1}-V_t \ge \lambda c_j$, equivalently $(V_{t+1}-V_t)/c_j \ge \lambda$. $\square$

For exact sufficiency, the non-cost value is $v(A,G)=\mathbf{1}\{G\subseteq A\}$.
Adding next tool $j$ changes value by
\[
v(A\cup\{j\},G)-v(A,G)
=\mathbf{1}\{G\subseteq A\cup\{j\}\}-\mathbf{1}\{G\subseteq A\}.
\]
If $A$ is already sufficient this marginal is zero; if it is not, the marginal is one exactly when $G\setminus A=\{j\}$.
Thus the exact-payoff continuation value is about the probability that the next tool completes the remaining required set, not merely the probability that the next tool is individually relevant.

\subsection{Proof of Proposition~\ref{prop:index}}

Under the additive model the expected payoff of any $A$ is
\begin{equation*}
\mathbb{E}[U(A,x)\mid x] = \sum_{j\in A}\bigl(p_j(x) - \lambda c_j\bigr),
\end{equation*}
a sum of per-tool terms that do not interact.
The sum is maximized by including exactly the tools with nonnegative terms: $j\in A^{*}(x)$ iff $p_j(x)-\lambda c_j\ge0$, i.e.\ $p_j(x)/c_j\ge\lambda$.

For the fixed-ordering parts: restricted to the score ordering, the only feasible sets are prefixes, and the prefix of depth $t$ has expected payoff $\sum_{r=1}^{t}\bigl(p_{(r)}(x)-\lambda c_{(r)}\bigr)$ by the same additivity, so the optimal depth maximizes this partial sum.
Continuing from depth $t$ to depth $k>t$ changes the expected payoff by $\sum_{r=t+1}^{k}\bigl(p_{(r)}(x)-\lambda c_{(r)}\bigr)$; some continuation is strictly preferred to stopping at $t$ iff the maximum of these increments over $k>t$ is strictly positive, which is the stated condition. $\square$

\subsection{Proof of Proposition~\ref{prop:fail}}

By assumption $p_i>p_j$, so any rule that ranks or thresholds on value scores alone admits $i$ whenever it admits $j$.
But $p_j/c_j > p_i/c_i$, so by Proposition~\ref{prop:index} there exist cost pressures satisfying $p_i/c_i < \lambda < p_j/c_j$ at which the optimal rule includes $j$ and excludes $i$ (the open interval avoids the boundary ties at either endpoint, where the optimum is not unique).
On this range a rule that ranks or thresholds on value scores alone faces three cases, all of which disagree with the optimal cost-aware rule $\{j\}$: a threshold $\tau\le p_j$ (or top-$2$ ranking) includes both, over-acquiring $i$; a threshold satisfying $p_j<\tau\le p_i$ (or top-$1$ ranking) includes only $i$, the exact inversion of the optimum; and a threshold $\tau>p_i$ excludes both, missing $j$. $\square$

A concrete two-tool instance makes the inversion visible.
Let $p_i=0.8$, $c_i=4$, $p_j=0.5$, $c_j=1$, and $\lambda=0.3$.
The score order prefers $i$ because $0.8>0.5$.
The cost-aware surplus terms are $p_i-\lambda c_i=-0.4$ and $p_j-\lambda c_j=+0.2$, equivalently $p_i/c_i=0.2<\lambda<p_j/c_j=0.5$.
The optimal cost-aware subset is therefore $\{j\}$, while any top-1 score rule chooses the wrong tool.

\begin{table*}[t]
\centering
\scriptsize
\setlength{\tabcolsep}{4pt}
\begin{tabular}{p{0.17\textwidth}p{0.21\textwidth}p{0.21\textwidth}p{0.11\textwidth}p{0.19\textwidth}}
\toprule
Method & Decision object & Cost model & Model access & Decision policy \\
\midrule
Adaptive-RAG \citep{jeong2024adaptiverag} & how much retrieval per query & none explicit & black-box & predict-then-threshold \\
DynamicRAG \citep{sun2025dynamicrag} & reranking depth per query & implicit (fewer docs) & black-box & RL from answer feedback \\
Stop-RAG \citep{park2025stoprag} & stop iterative RAG rounds & none explicit & black-box & value-based $Q(\lambda)$ \\
EcoAct \citep{zhang2024ecoact} & when to register a tool & token cost, prompted & black-box & prompted, not learned \\
Hidden-state probes \citep{wu2026tocall,sun2026when2tool} & single call vs.\ no-call & per-call & internal states & learned probes \\
Utility-guided \citep{liu2026utilityguided} & respond/retrieve/tool/stop & hand-set utility & black-box & hand-designed rule \\
Ranked-list truncation \citep{bahri2020choppy,meng2024rlt} & cut a ranked document list & uniform effort & black-box & learned, score-only \\
\textbf{CAM-DF (this paper)} & stop along a tool ranking & explicit $\mathrm{suff}-\lambda\,\mathrm{cost}$, heterogeneous $c_j$ & black-box & regret-weighted stopping classifier \\
\bottomrule
\end{tabular}
\caption{Positioning map: closest neighbors versus CAM-DF's black-box ranked-prefix acquisition layer. Every cell consolidates a distinction already stated in the main-text Related Work; no new claims.}%
\label{tab:positioning}%
\end{table*}

\section{Experimental Details}%
\label{app:setup}%

\paragraph{Retail cost vector.}
Table~\ref{tab:costs} lists the base costs $\bar c$ before mean-normalization.
This author-specified sensitivity vector is intended to encode that profile, order, and product lookups are operationally heavier than utility calls; it is not measured production telemetry.
Dispersion-$d$ costs are $c_j(d) = \max\{0.10,\; 1 + d(\bar c_j/\bar c_{\mathrm{mean}} - 1)\}$.

\begin{table}[t]
\centering
\small
\begin{tabular}{lc}
\toprule
Tool & Base cost $\bar c_j$ \\
\midrule
calculate & 0.50 \\
find\_user\_id\_by\_email & 0.70 \\
find\_user\_id\_by\_name\_zip & 0.80 \\
list\_all\_product\_types & 1.00 \\
get\_user\_details & 1.10 \\
get\_order\_details & 1.80 \\
get\_product\_details & 2.00 \\
\bottomrule
\end{tabular}
\caption{Retail base tool costs before mean-normalization.}%
\label{tab:costs}%
\end{table}

\begin{table*}[t]
\centering
\scriptsize
\setlength{\tabcolsep}{3pt}
\renewcommand{\arraystretch}{1.06}
\begin{tabular}{p{0.20\textwidth}p{0.73\textwidth}}
\toprule
Component & Setting \\
\midrule
Retail benchmark & 67 $\tau$-bench Retail tasks; 7 read-only candidate tools; required-tool sets $G_x$ from public annotations; 30 random 55/20/25\% train/validation/test splits ($\approx$36/13/18 tasks). \\
Ranking instantiations & Five frozen LLM ranking sources: qwen-plus (main), qwen-plus conservative prompt, qwen-turbo, qwen-max, and deepseek-v3. Each policy comparison is paired within the same cached ranking source. \\
Retail payoff grid & Exact payoff in Eq.~\ref{eq:uexact}; $\lambda\in\{0.05,0.12,0.20\}$; cost dispersion $d\in\{0,1.0,1.5\}$ with $c_j(d)=\max\{0.10,1+d(\bar c_j/\bar c_{\mathrm{mean}}-1)\}$. \\
Learners & Standardized public features; $\ell_2$-regularized logistic
regression for CAM-DF, CAM-DF-lite, Aggregate-only DF, and
predict-then-threshold stopping. The implementation parameterizes the
regularization term $\gamma$ using the inverse regularization strength
$C=1.0$ (liblinear; maximum 400 iterations in the Retail main pipeline
and 500 in the robustness and Atlas pipelines). \\
Validation tuning & Fixed-$k$, score threshold, score-per-cost threshold, Retail-rule top-up threshold, predict-then-threshold stopping threshold, and the CAM-DF-lite stopping thresholds (both Retail and MCP-Atlas) are tuned on validation payoff only; grids use 19 score/probability thresholds, 30 Retail score-per-cost thresholds, and 25 MCP-Atlas thresholds. Full CAM-DF and Aggregate-only DF use the fixed $0.5$ stop gate (Table~\ref{tab:tuning} shows tuning CAM-DF's threshold does not help it). \\
Uncertainty & Retail paired task bootstrap averages each task over its test appearances and resamples 67 tasks 2{,}000 times; split SEs report variation over the 30 splits. \\
MCP-Atlas & 495 usable multi-tool tasks: of the first 500 public dataset rows, those with a nonempty enabled-tool list and at least one logged trajectory tool among the enabled candidates (all 495 have $\ge2$ required tools); 20 random 60/20/20\% splits; partial payoff; unit tool costs; qwen-plus main and qwen-turbo replication; paired bootstraps use the 485 tasks with complete policy pairs. \\
MCP-Atlas heterogeneous costs & Ten deterministic pseudo-random cost vectors (md5 of ``\{vector\}:\{tool\}'' $\to$ base $\in[0.5,2.0]$, mean-normalized, dispersion $d=1.0$); 10 seeds per vector; partial payoff; adds score-per-cost threshold, cost-aware lite features, and a cost-blind lite control. \\
Dense Retail grid & $\lambda\in\{0.05,\dots,0.20\}\times d\in\{0,\dots,2.0\}$ ($5\times5$), 30 splits per cell, identical conservative pipeline; the nine main-grid cells reproduce released values exactly. \\
MetaTool reduced probe & 497 public multi-tool tasks; 15 candidate tools appearing as required in the subset; no live execution, no native costs, and exact two-tool required labels only. We use an unsupervised train-split TF-IDF cosine scorer and unit costs, so this is a boundary check rather than main evidence. \\
Live Retail execution & qwen-plus agent and user simulator, temperature 0, up to 30 steps; read tools restricted to each acquired set while write/action tools remain available; one main setting ($\lambda=0.12,d=1.0$); all five conditions replicated with an independent qwen-max user simulator (same agent and acquired sets). \\
\bottomrule
\end{tabular}
\caption{Implementation and tuning settings. This table is the appendix-level audit trail for experimental setup and hyperparameters; code and cached artifacts follow the same split/tuning protocol.}%
\label{tab:settings}%
\end{table*}

\paragraph{Splits and tuning.}
Table~\ref{tab:settings} summarizes the split, tuning, hyperparameter, and bootstrap protocol.
All thresholds and $k$ are tuned on validation payoff for each $(\lambda,d,\text{seed})$; CAM-DF and all learned baselines are refit per split.

\paragraph{Pipeline reconciliation map.}
Nominally identical comparisons can differ by small amounts across the released pipelines (for example, predict-then-threshold payoff $0.352$ in the main table versus $0.361$ in the $\Delta$-regression rerun); every table caption names its pipeline, and all inference is within-pipeline paired.
Table~\ref{tab:pipelines} maps each pipeline to the tables it feeds and to the features its predict-then-threshold baseline observes.

\begin{table*}[t]
\centering
\scriptsize
\setlength{\tabcolsep}{4pt}
\begin{tabular}{>{\raggedright\arraybackslash}p{0.20\textwidth}
                >{\raggedright\arraybackslash}p{0.33\textwidth}
                >{\raggedright\arraybackslash}p{0.20\textwidth}
                >{\raggedright\arraybackslash}p{0.19\textwidth}}
\toprule
Pipeline & Feeds & Prediction-baseline features & Notes \\
\midrule
Conservative lite-comparison & Main Table~1 (Retail block); absolute Retail table (Table~\ref{tab:retailabs}); dense grid (Table~\ref{tab:griddense}); feature-matched control (Table~\ref{tab:costaware}); lite coefficients (Table~\ref{tab:coef}) & 16 cost-inclusive public aggregates & CAM-DF/Aggregate-only DF use the fixed $0.5$ gate; lite and all heuristics validation-tuned \\
Released grid/cross-scorer & Full $\lambda$ grid (Table~\ref{tab:grid}); cross-scorer (Table~\ref{tab:cross}); ablations (Table~\ref{tab:ablation}) & 9 score-only aggregates (no cost features) & source of the released 10/10 cells; the feature-matched control above removes the cost-visibility confound \\
Robustness reruns & ordering, $\Delta$-regression, partial payoff, residual correction, tuning symmetry (Secs.~\ref{sec:exp-ablation}--\ref{sec:exp-robust}) & refit within each experiment; numbers labeled ``rerun'' & within-run controls; not cross-table comparable \\
MCP-Atlas unit-cost & Table~\ref{tab:atlas}, Fig.~\ref{fig:atlas} & 15 cost-free lite features (ridge under partial payoff) & unit costs; CAM-DF-lite only \\
MCP-Atlas heterogeneous-cost & Table~\ref{tab:atlashet} & cost-aware lite features (ridge under partial payoff) & ten random cost vectors; adds a cost-blind lite control \\
\bottomrule
\end{tabular}
\caption{Pipeline reconciliation map: which pipeline feeds which table and what the predict-then-threshold baseline observes in each.}%
\label{tab:pipelines}%
\end{table*}

\paragraph{MCP-Atlas protocol.}
Per seed, the sufficiency predictor (a ridge regressor under the partial payoff) and the CAM-DF-lite stopping classifier (regret-weighted logistic regression on the cost-free lite features) are refit on the training split.
Candidates are each task's listed tools (mean $15.21$, max $37$), with $c_j\equiv1$; paired bootstraps use the 485 tasks with complete policy pairs across all baselines.

\paragraph{Reproducibility.}
All cached LLM scores (five Retail ranking configurations used to instantiate the interface; MCP-Atlas qwen-plus and qwen-turbo scores), cost vectors, seeds, and evaluation scripts are included in the anonymized release artifact.
Every offline replay number in the paper is generated by the released pipeline; live execution numbers trace to released non-dry logs, with reruns requiring tau-bench setup and LLM/API credentials.

\paragraph{Using the training recipe outside replay.}
Training needs required-set labels and a cost pressure.
For prospective use, $G_x$ could be mined from logged successful trajectories (the tools consulted before a verified correct action, our MCP-Atlas proxy) or labeled offline by judges, but trajectory-mined sets are noisy proxies that require audit or judge validation rather than automatic ground truth.
The supervision-noise experiment shows the learned edge surviving roughly $5\%$ membership noise and inverting by $10\%$, so verified rather than heuristic labels are the binding input.
The cost pressure $\lambda$ prices tool cost in units of task success.
Operators can choose or sweep it against validation budget or latency constraints; Table~\ref{tab:grid} reports the full grid.

\section{Algorithmic and Prompt Details}%
\label{app:algo}%

This section spells out the implementation-facing details behind the CAM-DF training/inference recipe of main-text Section~5.
It follows the same disclosure role as an appendix algorithm/prompt section: enough detail to reconstruct the replay target, the deployed gate, and the scoring interface without relying on prose interpretation.

\begin{figure*}[t]
\refstepcounter{algorithm}\label{alg:frontier}
\small
\noindent\rule{\textwidth}{0.4pt}
\vspace{-0.8ex}
\noindent\textbf{Algorithm \thealgorithm: Offline prefix-frontier target construction.}\par
\vspace{0.8ex}
\begin{tabular}{@{}r >{\raggedright\arraybackslash}p{0.87\textwidth}@{}}
\textbf{Input} & Task $x$, ordered candidates $(j_1,\ldots,j_m)$, costs $c_j$, cost pressure $\lambda$, offline label set $G_x$, payoff form $U(A,x)$. \\
1 & Build nested prefixes $A_0=\emptyset$ and $A_t=\{j_1,\ldots,j_t\}$ for $t=1,\ldots,m$. \\
2 & For every prefix, compute $C_t=\sum_{j\in A_t}c_j$ and $U_t=U(A_t,x)$; for the main exact payoff, $U_t=\mathbf{1}\{G_x\subseteq A_t\}-\lambda C_t$. \\
3 & Compute the backward frontier $F_m=U_m$ and $F_t=\max\{U_t,F_{t+1}\}$ for $t=m-1,\ldots,0$. \\
4 & For each nonterminal $t<m$, set $Q_{\mathrm{stop}}(t)=U_t$ and $Q_{\mathrm{cont}}(t)=F_{t+1}$. \\
5 & Set $\Delta_t=Q_{\mathrm{stop}}(t)-Q_{\mathrm{cont}}(t)$, $y_t=\mathbf{1}\{\Delta_t\ge0\}$, and $w_t=|\Delta_t|+\epsilon$ with $\epsilon=10^{-4}$ in code to keep zero-gap examples numerically present. \\
6 & Emit training row $(\phi_t,y_t,w_t)$ where $\phi_t$ contains only public score, cost, order, and prefix-progress features; never include $G_x$, $|G_x|$, or missing-required indicators. \\
\end{tabular}
\vspace{-0.6ex}
\noindent\rule{\textwidth}{0.4pt}
\vspace{-1.2ex}
\end{figure*}

\begin{figure*}[t]
\refstepcounter{algorithm}\label{alg:deploy}
\small
\noindent\rule{\textwidth}{0.4pt}
\vspace{-0.8ex}
\noindent\textbf{Algorithm \thealgorithm: Runtime pre-execution CAM-DF gate.}\par
\vspace{0.8ex}
\begin{tabular}{@{}r >{\raggedright\arraybackslash}p{0.87\textwidth}@{}}
\textbf{Input} & User task, candidate order or scores from the existing harness, costs, cost pressure $\lambda$, learned gate $\pi_\theta$, feature map $\phi$. \\
1 & Initialize planned acquisition prefix $A_0=\emptyset$. No candidate tool has executed yet. \\
2 & For $t=0,\ldots,m-1$, form the public state $\phi_t=\phi(x,A_t,s,c,\lambda)$. \\
3 & If $\pi_\theta(\phi_t)\ge0.5$, stop and return prefix $A_t$ to the harness (CAM-DF-lite is fit in the mirror-image continue orientation: it scores $p_{\mathrm{cont}}=\Pr(\Delta<0)$ and stops when $p_{\mathrm{cont}}$ falls below its validation-tuned threshold). \\
4 & Otherwise append the next ranked candidate to the planned prefix, $A_{t+1}=A_t\cup\{j_{t+1}\}$, and continue the virtual prefix walk. \\
5 & If no earlier state stops, return $A_m$. The harness then executes or exposes only this selected prefix before the downstream answer/action. \\
\textbf{Boundary} & A continue decision in this algorithm is not a real tool call and does not reveal tool output. Output-adaptive next-tool choice after observing a tool result is a future multi-round extension, not the current replay abstraction. \\
\end{tabular}
\vspace{-0.6ex}
\noindent\rule{\textwidth}{0.4pt}
\vspace{-1.2ex}
\end{figure*}

\subsection{Feature Inventory}

The full CAM-DF feature map contains 39 public features in four groups.
The aggregate selected/remaining block records prefix progress, selected and remaining score summaries, selected and remaining cost summaries, and selected/remaining score-per-cost totals.
The marginal block records the next candidate's score, cost, score/cost ratio, high-cost indicator, and related score-cost interactions.
The cost-pressure block records dispersion and $\lambda$-scaled next-cost effects.
The final optional identity block is a one-hot indicator for the next candidate tool; it is used only in full CAM-DF and is removed in CAM-DF-lite to reduce dependence on a fixed catalog.

CAM-DF-lite uses exactly ten features:
progress $t/m$; next score; next score divided by next cost; next-score gap to the second remaining candidate; remaining maximum score; normalized remaining-score sum; remaining score divided by remaining cost; normalized selected cost; $\lambda$ times next cost; and the next-high-cost indicator.
These are the features behind Table~\ref{tab:coef}.
They deliberately omit tool identity, offline labels, required-set size, and any outcome-derived information.

\subsection{Retail Scoring Prompt Template}

The prompt below is the default replay scorer template used to generate the qwen-style cached Retail necessity/value scores.
It is an experimental way to populate the candidate-order interface; CAM-DF itself only requires a candidate order or score signal from some upstream harness component.

\begin{quote}
\small
\textbf{System/user prompt template.}
You are helping design a retail customer-service agent.
The agent must decide which read-only information tools to call before taking any consequential business action such as cancel, return, exchange, or modify.
Score each candidate tool from 0 to 1 for how necessary/useful it is for this specific user request.
High score means the agent likely needs this tool before it can safely decide; low score means probably unnecessary.
Do not score final mutation/action tools because they are not candidates here.

\emph{User request:} [task id and user instruction]

\emph{Candidate read-only tools:} [tool name and natural-language description for each candidate]

Return only a JSON object whose keys are exactly the tool names and whose values are numbers between 0 and 1. No explanation.
\end{quote}

The conservative-prompt variant keeps the same input/output contract but asks for high scores only when a read-only tool is clearly needed before an irreversible business action.
The score-generation temperature is $0$ and cached JSONL rows store the task id, required set, and scores.
Policy training and evaluation read those caches; no LLM call is needed to reproduce offline replay results.

\subsection{Artifact and Implementation Trace}

The implementation mirrors Algorithms~\ref{alg:frontier}--\ref{alg:deploy}.
Retail score caches live in the structured-tool-acquisition pilot results directory; selected older scorer caches are kept in that pilot's legacy-results archive.
CAM-DF, CAM-DF-lite, cross-scorer, ablation, tuning, label-noise, MCP-Atlas, MetaTool reduced-probe, and live-policy analysis scripts live in the tool-acquisition-correction scripts directory.
The results digest is regenerated from frozen CSV/JSONL artifacts and is the single source of truth for manuscript numerals.

\section{Additional Robustness Checks}%
\label{app:robust}%
This appendix expands the five checks summarized in Section~\ref{sec:exp-ablation}; all use the Retail benchmark.

\paragraph{Score order vs.\ $s_j/c_j$ order.}
Proposition~\ref{prop:fail} indicts score-only orderings, yet CAM-DF acquires in score order, so we re-ran CAM-DF with acquisition re-ordered by $s_j/c_j$ (same labels, features, and loss).
The orderings genuinely differ, selecting different prefixes on $65.6\%$ of $\lambda=0.12$ test tasks, but payoffs are statistically indistinguishable: paired differences (score minus $s_j/c_j$ order) at $\lambda=0.12$ are $+0.000$, $+0.003$, $+0.007$ for $d=0,1.0,1.5$, all CIs covering zero (at $d=0$ the orderings coincide).
Table~\ref{tab:order} reports the full $\lambda\times d$ grid: no cell significantly favors the $s_j/c_j$ order (the most negative mean, $-0.015$ at $\lambda=0.05$, $d=1.5$, has CI $[-0.043, +0.003]$), and at $\lambda=0.20$, $d=1.5$ the score order is significantly better ($+0.053$, CI $[+0.002, +0.098]$, task win rate $0.84$).
Re-ordering can even lower the attainable ceiling: the $s_j/c_j$-order oracle payoff at $d=1.5$ is $0.385$ versus $0.419$ under score order.
In this Retail score-quality regime, cost awareness is carried mostly by the stopping layer, which absorbs much of the ordering difference; where dispersion and cost pressure are largest, the upstream ranking's required-set signal is worth more than the $s_j/c_j$ re-weighting that scrambles it.

\paragraph{$\Delta$-regression target.}
To distinguish direct stopping-gap regression from regret-weighted classification, we regress $\Delta(S_t)$ on $X_t$ and stop when the prediction is nonnegative.
This performs on par with CAM-DF: paired differences are $+0.025$, $+0.006$, and $-0.006$ at $\lambda=0.12$, all with CIs covering zero.
Like CAM-DF, it sits above predict-then-threshold stopping ($0.372/0.319/0.297$ vs.\ $0.361/0.273/0.235$; both rerun within this experiment's pipeline).
We read this honestly: the gain comes from targeting the decision object $\Delta$ itself; the classification form adds the alignment of Theorem~\ref{thm:bayes} and a slight edge under uniform cost.

\paragraph{Cost-vector perturbation.}
The heterogeneous-cost results rest on an author-specified base cost vector, so we re-ran $\lambda=0.12$, $d=1.0$ under 20 random permutations of the base costs across the seven tools (10 seeds each).
CAM-DF beats predict-then-threshold stopping in 20 of 20 permutations (mean $+0.059$, range $+0.027$ to $+0.110$), with the per-permutation 95\% CI excluding zero from above in 17 of 20 and never excluding zero from below; against the score-per-cost threshold the mean is $+0.015$, positive in only $65\%$ of permutations (negative mean in 7 of 20), with the CI excluding zero in just 2 of 20.
The advantage over the learned baseline is robust to which tools happen to be expensive; against the tuned score-per-cost heuristic, what is robust is the main-setting statistical tie.

\paragraph{Tuning-protocol controls.}
Baseline thresholds are tuned on a ${\approx}$13-task validation split while CAM-DF uses a fixed $0.5$ stop threshold, so part of the reported gaps could in principle be validation tuning noise.
Symmetrizing the protocol either way says otherwise (Table~\ref{tab:tuning}).
Giving CAM-DF a validation-tuned threshold does not help it: the tuned thresholds drift below the fixed $0.5$ (per-$d$ medians $0.32$/$0.25$/$0.25$) yet leave payoff unchanged.
The differences for CAM-DF minus its tuned variant are $+0.016$/$+0.018$/$+0.007$ across $d$, all with CIs covering zero, and payoff varies by less than $0.01$ across fixed thresholds $0.3$--$0.5$ at every dispersion.
Denying the prediction baseline its tuning instead widens the gap to $+0.070$--$+0.083$ (CIs strictly positive).
With both policies validation-tuned, the gap remains strictly positive ($+0.030$/$+0.044$/$+0.060$).

\paragraph{Training-free stopping predicates.}
Training-free stopping rules \citep{kieback2026tasr} argue for fixed predicates with no learned controller, so we evaluate four predicates whose parameters are fixed a priori and never see the train or validation split:
(i) acquire every tool with $s_j \ge 0.5$;
(ii) cut the score-descending ranking at the largest adjacent score drop;
(iii) take the smallest prefix holding $80\%$ of the total score mass; and
(iv) the naive Proposition~\ref{prop:index} plug-in acquiring $j$ iff $s_j \ge \lambda c_j$, which treats the score as a calibrated marginal success probability.
Predicates are evaluated on the same per-seed test splits as CAM-DF, so the paired task bootstrap is like-for-like (Table~\ref{tab:trainfree}).
CAM-DF beats the plug-in in 7 of 9 cells; the failure mechanism is systematic over-acquisition (sufficiency $0.97$--$0.98$ at $5.3$--$6.4$ tools versus CAM-DF's $3.2$--$5.2$), because raw scores are not calibrated marginal payoffs.
At $\lambda=0.20$, $d=1.5$ every predicate loses significantly ($+0.100$ to $+0.212$).
The strongest predicate, the fixed $0.5$ threshold, is a statistical tie at $\lambda=0.12$ ($+0.022$ to $+0.034$, CIs covering zero), consistent with the main-text boundary that simple rules are competitive at moderate pressure, while the predicates' cost- and $\lambda$-blindness costs them exactly where cost pressure and heterogeneity bind.

\paragraph{Residual-corrected prediction (negative result).}
If the gap were merely cost-related miscalibration of the sufficiency predictor, a double-machine-learning residual correction \citep{chernozhukov2018dml} should close it: we cross-fit a ridge model of the sufficiency residuals, corrected the prediction, and re-tuned the threshold.
The diagnostic finds modest cost-related structure (mid-cost states over-predicted), but the corrected policy's payoff is statistically indistinguishable from the uncorrected one ($0.366$ vs.\ $0.366$ at $d=0$; $0.289$ vs.\ $0.285$ at $d=1.0$; $0.247$ vs.\ $0.247$ at $d=1.5$) and stays below the same-pipeline decision-focused policy (CAM-DF rerun within this experiment: $0.370$, $0.290$, $0.269$; SEs $\le 0.015$) at every dispersion.
Fixing the prediction is not fixing the decision.

\paragraph{LLM self-selection reference.}%
\label{app:selfselect}%
As a descriptive reference (one prompt, all 67 tasks, no splits, not a tuned competitor), we let qwen-plus itself choose the tool set given the task, tools, costs, and $\lambda=0.12$.
It under-acquires substantially (sufficiency $0.39$/$0.31$ at $d=0$/$1.0$, payoffs $+0.024$/$-0.121$, versus $0.400$/$0.331$ for CAM-DF on the same scores), consistent with cost-centric agent evaluations \citep{liu2025budgetaware,liu2025costbench}: the LLM does not self-manage acquisition under cost pressure.

\section{End-to-End Execution on Retail}%
\label{app:e2e}%
The main payoff scores an acquired set by the sufficiency indicator $\mathbf{1}\{G_x\subseteq A\}$ rather than by running the agent. To check that this proxy tracks realized task success, we execute acquired sets in the live $\tau$-bench Retail benchmark: qwen-plus serves as both the tool-calling agent and the user simulator (temperature $0$, up to $30$ steps), and tau-bench scores each episode $0/1$ by checking the resulting database state. The agent's \emph{read-only} tools are restricted to a set $A$ while all write/action tools remain available, so the experiment isolates the effect of read-tool acquisition on real success. We compare five per-task conditions over all $67$ tasks (Table~\ref{tab:e2e}): \emph{full} ($A$ = all seven read candidates), \emph{oracle} ($A=G_x$), \emph{insufficient} ($A=G_x$ minus its lowest-scored required tool), and the CAM-DF/CAM-DF-lite acquired sets from the main $\lambda=0.12,d=1.0$ policy configuration.

\begin{table*}[t]
\centering
\small
\begin{tabular}{lcccc}
\toprule
Condition & Success (95\% CI) & Succ. & Read tools & Calls \\
\midrule
full ($A=$ all 7)        & $0.58$ $[0.46,0.69]$ & 39/67 & 7.00 & 6.8 \\
oracle ($A=G_x$)         & $0.67$ $[0.55,0.77]$ & 45/67 & 3.73 & 6.6 \\
insufficient ($A=G_x{-}1$) & $0.24$ $[0.15,0.35]$ & 16/67 & 2.73 & 5.0 \\
CAM-DF policy            & $0.67$ $[0.55,0.77]$ & 45/67 & 4.39 & 7.0 \\
CAM-DF-lite policy       & $0.64$ $[0.52,0.75]$ & 43/67 & 4.31 & 6.6 \\
\bottomrule
\end{tabular}
\caption{End-to-end tau-bench Retail success (Wilson 95\% CIs; mean read tools exposed and mean tool calls per episode) under read-tool restriction. Realized success tracks the sufficiency proxy: $A=G_x$ matches the full-tool agent (paired McNemar $p=0.21$) while dropping one required tool collapses success ($p<0.001$). The CAM-DF and CAM-DF-lite policy-acquired sets also beat insufficient acquisition ($p<0.001$), with no detectable difference from full/oracle access in this underpowered 67-task self-play check. Pooled across the original proxy conditions, proxy-sufficient acquisitions ($A\supseteq G_x$) succeed $0.63$ $[0.54,0.70]$ ($84/134$) versus $0.24$ $[0.15,0.35]$ for proxy-insufficient ones.}%
\label{tab:e2e}%
\end{table*}

Realized success tracks the proxy. Giving the agent \emph{only} the required read tools (oracle) shows no detectable benefit from the extra four tools relative to all seven (full): $0.67$ vs $0.58$, a paired McNemar test on the shared 67 tasks not rejecting equality ($p=0.21$; if anything favoring oracle, oracle-only-win $11$ vs full-only-win $5$, plausibly because fewer distractor tools focus the agent; this is an underpowered null at $n=67$, not a proven equivalence). Dropping a single required tool collapses success from $0.67$ to $0.24$ (oracle wins $34$ tasks, insufficient $5$; $p<0.001$), even though the insufficient agent issues \emph{fewer} tool calls ($5.0$ vs $6.6$): under-acquisition is cheaper but fails. The proxy is not conservative: the dropped (lowest-scored) required tool is load-bearing for real success. It is also not perfect (insufficient still succeeds on $24\%$ of tasks, where the agent works around the missing tool or the used-tools annotation slightly overstates the minimal set), but the gap between proxy-sufficient and proxy-insufficient acquisitions ($0.63$ vs $0.24$) supports $\mathbf{1}\{G_x\subseteq A\}$ as a useful end-to-end success proxy and suggests that $G_x$ coverage is often sufficient in this harness. The learned policy rows close the dry-run boundary: CAM-DF succeeds on $0.67$ (45/67) and CAM-DF-lite on $0.64$ (43/67), both far above insufficient acquisition (CAM-DF policy-only wins $29$ tasks vs $0$ insufficient-only; CAM-DF-lite $30$ vs $3$; exact $p<0.001$). Task-level diagnostics support the same bounded reading rather than a tautological proxy story: CAM-DF acquires exactly $G_x$ on 29 tasks, a strict superset on 34, and misses at least one annotated required read tool on 4; its proxy/live cells are 41 covered successes, 22 covered failures, 4 missing-but-successful tasks, and 0 missing failures. CAM-DF-lite gives the analogous 28/31/8 exact/superset/missing split, with 41 covered successes, 18 covered failures, 2 missing-but-successful tasks, and 6 missing failures. Mean tool calls do not explain success monotonically (CAM-DF success/failure calls $6.89/7.32$; CAM-DF-lite $6.65/6.50$), so the live check is about information sufficiency rather than simply calling more tools.
The acquisition-side saving is visible in exposure: the policies provision $4.39$/$4.31$ of the seven read tools on average (full access provisions all $7$) at statistically indistinguishable success: exposure is the cost object the stopping layer controls, while realized call counts are the downstream agent's choice and stay comparable. The policy rows do not prove live dominance over full/oracle access: CAM-DF vs full has exact $p=0.210$, CAM-DF vs oracle $p=1.000$, CAM-DF-lite vs full $p=0.424$, and CAM-DF-lite vs oracle $p=0.804$.

\paragraph{Independent user-simulator replication.}
Because qwen-plus serves as both agent and user simulator above (self-play, temperature $0$), we reran all five conditions with qwen-max as the user simulator, holding the agent model and every acquired set fixed (Table~\ref{tab:e2eusim}).
The pattern replicates: insufficient acquisition collapses to $0.21$ (14/67), the learned policies reach $0.63$ (CAM-DF, 42/67) and $0.64$ (CAM-DF-lite, 43/67), and full/oracle sit at $0.66$ (44/67).
Paired exact tests under the independent simulator: CAM-DF beats insufficient 29-policy-only-wins to 1 ($p<0.001$; CAM-DF-lite 29 to 0), while policies remain statistically indistinguishable from full/oracle (CAM-DF vs full $p=0.727$, vs oracle $p=0.727$; CAM-DF-lite $p=1.000$ for both).
One simulator-dependent detail is worth noting honestly: under qwen-max the full-tool condition improves from $0.58$ to $0.66$, so the self-play run's suggestion that distractor tools hurt the full agent does not replicate; none of our claims rests on that comparison.
The remaining live caveats are the single scorer/cost setting and the ranked-prefix (not online multi-round) scope.

\begin{table}[t]
\centering
\footnotesize
\setlength{\tabcolsep}{3.5pt}
\begin{tabular}{lcc}
\toprule
Condition & Self-play & Indep.\ qwen-max sim \\
\midrule
full ($A=$ all 7)          & $0.58$ (39/67) & $0.66$ (44/67) \\
oracle ($A=G_x$)           & $0.67$ (45/67) & $0.66$ (44/67) \\
insufficient ($A=G_x{-}1$) & $0.24$ (16/67) & $0.21$ (14/67) \\
CAM-DF policy              & $0.67$ (45/67) & $0.63$ (42/67) \\
CAM-DF-lite policy         & $0.64$ (43/67) & $0.64$ (43/67) \\
\bottomrule
\end{tabular}
\caption{Live $\tau$-bench Retail success under two user simulators (agent qwen-plus, temperature $0$, identical acquired sets per condition). The proxy-calibration pattern (policies and oracle far above insufficient acquisition, indistinguishable from full access) replicates under the independent qwen-max simulator.}%
\label{tab:e2eusim}%
\end{table}

The caveats are therefore narrower but still important: the policy check covers one scorer/cost setting, and the experiment is still ranked-prefix acquisition rather than online multi-round control.

\section{Cross-Domain Scale and Replication Details}%
\label{app:crossdomain}%

This appendix expands the cross-domain scale test of Section~\ref{sec:exp-cross-domain}.
All three domains use the identical replay pipeline as Retail: candidates ordered by frozen qwen-plus necessity scores, exact and partial payoff, cost dispersions $d\in\{0,1.0,1.5\}$, $\lambda\in\{0.05,0.12,0.20\}$, 30 splits (50 for airline), and a task-level paired bootstrap.
The complete per-cell grids are in the released results digest; the tables below report the main cost pressure $\lambda=0.12$.

\paragraph{WorkBench (statistical-power domain).}
WorkBench \citep{styles2024workbench} is an outcome-centric agent benchmark whose grading compares the final sandbox database state to a per-task ground-truth outcome.
We use its 13 read-only information tools as the candidate pool (the 14 write tools are the consequential action, not candidates, matching the Retail/$\tau^2$ convention) and obtain 639 usable tasks after removing tasks with no read requirement.
Because the gold outcome records only write calls, the required-read set is not lifted from a gold read trajectory; it is the per-task majority consensus of the read calls issued by five strong reference agents whose write calls match the gold outcome (cross-model pairwise Jaccard $0.93$--$0.99$).
This is a \emph{derived} label, weaker than a human-read gold trajectory but stable across models; it is disclosed here and in the artifact, not silently treated as ground truth.
The scorer is strong on this domain (AUC $0.982$, cover-all rank $2.01$ of $13$, mean required-set size $1.64$).
Table~\ref{tab:workbench} shows the $\lambda=0.12$ result: against the tuned score-per-cost and score thresholds the paired gap covers zero under exact payoff at uniform cost (matching Retail) but excludes zero under heterogeneous cost and under partial payoff at every dispersion, with the effect small ($+0.006$ to $+0.014$).
The pattern holds across $\lambda$: at $\lambda=0.05$ the exact heterogeneous-cost score-per-cost cells are negative (the cost-blind ranking is well tuned at low pressure), and at $\lambda=0.20$ they are positive again, while the predict-then-threshold gap is positive at all three pressures under heterogeneous cost.

\begin{table}[t]
\centering
\small
\setlength{\tabcolsep}{3pt}
\begin{tabular}{llccc}
\toprule
Payoff & CAM-DF $-$ base & $d{=}0$ & $d{=}1.0$ & $d{=}1.5$ \\
\midrule
Exact & Score/cost thr. & $-0.004$ & $+0.014^{*}$ & $+0.008^{*}$ \\
      & Score thr. & $-0.004$ & $+0.031^{*}$ & $+0.040^{*}$ \\
      & Pred.-thr. & $+0.008^{*}$ & $+0.043^{*}$ & $+0.051^{*}$ \\
      & Agg.\ DF    & $0.000$ & $+0.036^{*}$ & $+0.042^{*}$ \\
\addlinespace
Partial & Score/cost thr. & $+0.007^{*}$ & $+0.008^{*}$ & $+0.006^{*}$ \\
        & Score thr. & $+0.007^{*}$ & $+0.013^{*}$ & $+0.018^{*}$ \\
        & Pred.-thr. & $+0.007^{*}$ & $+0.012^{*}$ & $+0.017^{*}$ \\
        & Agg.\ DF    & $0.000$ & $+0.007^{*}$ & $+0.014^{*}$ \\
\bottomrule
\end{tabular}
\caption{WorkBench ($n=639$, $\lambda=0.12$, full CAM-DF): paired-bootstrap payoff gap over each tuned baseline, per cost dispersion. Asterisk marks a 95\% CI excluding zero. The exact uniform-cost score-per-cost/score cells cover zero as on Retail, but heterogeneous-cost and all partial-payoff cells separate, at a small effect size. Aggregate-only DF is identically zero at $d{=}0$ (its cost features are constant there). Complete CIs and the $\lambda=0.05,0.20$ grid are in the results digest.}%
\label{tab:workbench}%
\end{table}

\paragraph{$\tau^2$-bench airline (low-power replication).}
$\tau^2$-bench \citep{barres2025tau2} is a dual-control tool-agent benchmark; we extract its airline domain (28 usable tasks, 6 read-only candidate tools, required mean $1.68$, qwen-plus scorer AUC $0.903$ / qwen-max $0.916$).
With only 28 tasks the CIs are wide, so this is a replication check, not a powered comparison.
CAM-DF significantly beats the non-learning heuristics that ignore cost structure: against acquiring all candidates the gap is $+0.228$ to $+0.255$ at $\lambda=0.12$ (all CIs positive), and against tuned fixed-$k$ it is significant at $\lambda=0.20$ under uniform and high-dispersion cost. Against the strongest learned baselines (predict-then-threshold stopping, Aggregate-only DF) and the tuned score-per-cost threshold the CIs cover zero, as expected at this sample size (Table~\ref{tab:tau2}).

\paragraph{$\tau^2$-bench telecom (dual-model proxy labels).}
The telecom domain (114 usable tasks, 21 read-only candidate tools, required mean $5.54$) is our largest-required-set domain.
To avoid a single model both labeling and scoring, required sets are \emph{shadow-labeled by qwen-max} and necessity scores are produced by qwen-plus; the reported AUC $0.943$ therefore measures two-model agreement, not recovery of a human gold set, and we read telecom as MCP-Atlas-level proxy evidence rather than a gold benchmark.
Telecom is an exact-payoff-hard domain: with required mean $5.54$ and $\lambda=0.20$, even the oracle often prefers acquiring nothing, so the exact-payoff comparisons degenerate.
The informative cells are heterogeneous-cost and partial payoff, where CAM-DF beats Aggregate-only DF, its cost-blind twin (partial $+0.020$ $[+0.015,+0.025]$ at $d=1.0$; $+0.022$ $[+0.017,+0.027]$ at $d=1.5$) and the tuned score-per-cost threshold under partial payoff at $d=1.0$ ($+0.036$ $[+0.027,+0.046]$); at uniform cost the tuned score-per-cost rule leads ($-0.021$ exact, $-0.013$ partial), the honest boundary for a hard set-coverage domain (Table~\ref{tab:tau2}).

\begin{table}[t]
\centering
\small
\setlength{\tabcolsep}{3pt}
\begin{tabular}{llccc}
\toprule
Domain & CAM-DF $-$ base & $d{=}0$ & $d{=}1.0$ & $d{=}1.5$ \\
\midrule
Airline & Acquire-all & $+0.255^{*}$ & $+0.228^{*}$ & $+0.234^{*}$ \\
(exact) & Fixed-$k$   & $+0.021$ & $+0.006$ & $+0.024$ \\
        & Score/cost thr.  & $+0.007$ & $-0.033$ & $-0.022$ \\
        & Pred.-thr.  & $+0.011$ & $-0.006$ & $+0.002$ \\
\addlinespace
Telecom & Agg.\ DF     & $0.000$ & $+0.014$ & $+0.008$ \\
(exact) & Score/cost thr.  & $-0.021^{*}$ & $+0.067^{*}$ & $+0.085^{*}$ \\
\addlinespace
Telecom & Agg.\ DF     & $0.000$ & $+0.020^{*}$ & $+0.022^{*}$ \\
(partial)& Score/cost thr. & $-0.013^{*}$ & $+0.036^{*}$ & $+0.010$ \\
\bottomrule
\end{tabular}
\caption{$\tau^2$-bench airline ($n=28$) and telecom ($n=114$) cross-domain replication ($\lambda=0.12$): paired-bootstrap payoff gap over each tuned baseline. Asterisk marks a 95\% CI excluding zero. Airline separates only from the cost-structure-ignoring heuristics (acquire-all); against strong learned baselines its 28-task CIs cover zero. Telecom is exact-payoff-hard: the cost-aware gain over the score-per-cost rule appears under heterogeneous cost, and the score-per-cost rule leads at uniform cost. Aggregate-only DF is identically zero at $d{=}0$ by construction. Full grids are in the results digest.}%
\label{tab:tau2}%
\end{table}

\FloatBarrier%
\section{Additional Tables}%
\label{app:tables}%

\begin{figure}[!b]
\centering
\includegraphics[width=0.7\columnwidth]{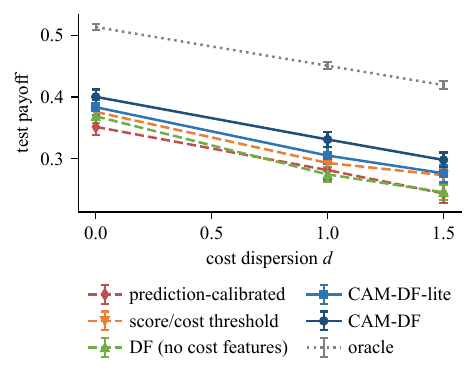}
\caption{Retail test payoff vs.\ cost dispersion $d$ ($\lambda=0.12$, qwen-plus scores; mean $\pm$ SE, 30 splits). CAM-DF has the best table mean at every $d$, CAM-DF-lite second, but the score-per-cost gaps at this $\lambda$ are not statistically separated. Its table-mean margin over predict-then-threshold stopping grows with $d$ ($0.048/0.050/0.054$); the margins over Aggregate-only DF ($0.031/0.056/0.052$) and the score-per-cost threshold ($0.024/0.038/0.025$) peak at $d=1.0$. Legend names: ``DF (no cost features)'' is Aggregate-only DF; ``score/cost threshold'' is the Score-per-cost threshold.}%
\label{fig:disp}%
\end{figure}

This section is grouped by the evidence questions used in the main paper.
First read the score and payoff-definition checks, then the robustness and ablation tables, then the external-validation and full-grid tables.
Figure~\ref{fig:disp} is a compact visual summary of the main Retail cost-dispersion pattern.
Table~\ref{tab:retailabs} adds SEs and the sufficiency/cost/waste decomposition behind the Retail columns of the main paper's Table~1; Table~\ref{tab:scores} gives the Retail score quality summary of Section~\ref{sec:exp-robust}; Tables~\ref{tab:cross}--\ref{tab:tuning} give the exact paired-bootstrap values for score-source robustness, ablations, and tuning controls; Table~\ref{tab:noise} gives the label-noise stress test; Table~\ref{tab:coef} gives the CAM-DF-lite coefficients; Table~\ref{tab:atlas} gives MCP-Atlas external validation; the short MetaTool paragraph below records a reduced boundary probe; Table~\ref{tab:partial} gives the payoff-definition robustness comparison of Section~\ref{sec:exp-robust}; Table~\ref{tab:grid} gives the full $\lambda$-grid bootstrap.

\subsection{Scores and Payoff Definitions}

\begin{table*}[t]
\centering
\small
\setlength{\tabcolsep}{5pt}
\begin{tabular}{lcccccc}
\toprule
& \multicolumn{3}{c}{Payoff (SE)} & \multicolumn{3}{c}{At $d{=}1.0$} \\
\cmidrule(lr){2-4}\cmidrule(lr){5-7}
Policy & $d{=}0$ & $d{=}1.0$ & $d{=}1.5$ & Suff. & Cost & Waste \\
\midrule
Tuned fixed-$k$        & $0.277\,{(0.014)}^{\dagger}$ & $0.219\,{(0.013)}^{\dagger}$ & $0.196\,{(0.012)}^{\dagger}$ & $0.885$ & $5.55$ & $1.31$ \\
Score threshold       & $0.376\,(0.012)$ & $0.302\,(0.012)$ & $0.267\,(0.012)$ & $0.924$ & $5.18$ & $0.90$ \\
Score-per-cost threshold       & $0.376\,(0.012)$ & $0.293\,(0.012)$ & $0.273\,(0.011)$ & $0.906$ & $5.10$ & $0.89$ \\
Retail rule           & $0.350\,(0.013)$ & $0.298\,(0.013)$ & $0.272\,(0.014)$ & $0.891$ & $4.94$ & $0.71$ \\
Predict-then-threshold & $0.352\,{(0.014)}^{\dagger}$ & $0.281\,{(0.015)}^{\dagger}$ & $0.244\,{(0.015)}^{\dagger}$ & $0.856$ & $4.78$ & $0.61$ \\
Aggregate-only DF           & $0.369\,(0.011)$ & $0.275\,{(0.013)}^{\dagger}$ & $0.246\,{(0.012)}^{\dagger}$ & $0.907$ & $5.27$ & $1.02$ \\
CAM-DF-lite           & $0.383\,(0.012)$ & $0.305\,(0.014)$ & $0.276\,(0.014)$ & $0.885$ & $4.84$ & $0.64$ \\
CAM-DF (full)         & $\mathbf{0.400}\,(0.012)$ & $\mathbf{0.331}\,(0.012)$ & $\mathbf{0.298}\,(0.012)$ & $0.922$ & $4.93$ & $0.68$ \\
\midrule
Prefix oracle         & $0.513\,(0.005)$ & $0.450\,(0.006)$ & $0.419\,(0.006)$ & $1.000$ & $4.58$ & $0.22$ \\
\bottomrule
\end{tabular}
\caption{Retail comparison behind the Retail columns of the main paper's Table~1, adding SEs and the sufficiency/cost/waste decomposition ($\lambda=0.12$, qwen-plus scores, 30 splits, conservative lite-comparison pipeline). Payoff mean by dispersion $d$ (SE over pooled task-split records; inference uses the paired task bootstrap); sufficiency/cost/waste at $d=1.0$. Bold: best non-oracle. $^{\dagger}$: CAM-DF's paired gain over this policy at this dispersion is strictly positive (95\% CI).}%
\label{tab:retailabs}%
\end{table*}

\begin{table}[t]
\centering
\small
\setlength{\tabcolsep}{4pt}
\begin{tabular}{lcccc}
\toprule
Scorer & Req. & Non-req. & AUC & Cover \\
\midrule
qwen-plus        & 0.867 & 0.239 & 0.946 & 4.04 \\
qwen-plus (cons.) & 0.771 & 0.102 & 0.894 & 4.45 \\
qwen-turbo       & 0.746 & 0.302 & 0.906 & 4.24 \\
qwen-max         & 0.791 & 0.194 & 0.953 & 3.97 \\
deepseek-v3      & 0.668 & 0.214 & 0.835 & 5.13 \\
\bottomrule
\end{tabular}
\caption{Retail tool-score quality by scorer (67 tasks): mean score of required and non-required tools, AUC, and the mean rank needed to cover all required tools (``Cover'').}%
\label{tab:scores}%
\end{table}

\begin{table}[t]
\centering
\small
\begin{tabular}{lcccc}
\toprule
 & \multicolumn{2}{c}{Exact (rob.)} & \multicolumn{2}{c}{Partial} \\
\cmidrule(lr){2-3}\cmidrule(lr){4-5}
Policy & $d{=}0$ & $d{=}1.0$ & $d{=}0$ & $d{=}1.0$ \\
\midrule
Pred.-thr.\ & 0.352 & 0.281 & 0.469 & 0.398 \\
Score-per-cost threshold  & 0.376 & 0.293 & 0.474 & 0.423 \\
CAM-DF-lite      & 0.378 & 0.307 & 0.478 & 0.421 \\
CAM-DF (full)    & \textbf{0.400} & \textbf{0.331} & \textbf{0.481} & \textbf{0.427} \\
\midrule
Prefix oracle    & 0.513 & 0.450 & 0.517 & 0.452 \\
\bottomrule
\end{tabular}
\caption{Retail payoff-definition robustness ($\lambda=0.12$, 30 splits): exact vs.\ partial-coverage payoff in the same robustness pipeline. Ordering is preserved and gaps shrink as the outcome becomes smoother. The exact columns are within-run controls for the partial-payoff rerun; the main exact comparison remains Table~\ref{tab:main}.}%
\label{tab:partial}%
\end{table}

\subsection{Robustness, Ablations, and Tuning}

\begin{table*}[t]
\centering
\small
\begin{tabular}{lccc}
\toprule
Scorer & AUC & $d=0$ & $d=1.0$ \\
\midrule
\keymetric{deepseek-v3}       & \keymetric{$0.835$} & \keymetric{$+0.103$ $[+0.062, +0.143]$} & \keymetric{$+0.095$ $[+0.052, +0.135]$} \\
qwen-plus (cons.) & 0.894 & $+0.081$ $[+0.026, +0.139]$ & $+0.067$ $[+0.025, +0.114]$ \\
qwen-turbo        & 0.906 & $+0.059$ $[+0.027, +0.094]$ & $+0.063$ $[+0.027, +0.104]$ \\
qwen-plus         & 0.946 & $+0.046$ $[+0.006, +0.089]$ & $+0.062$ $[+0.023, +0.104]$ \\
qwen-max          & 0.953 & $+0.065$ $[+0.030, +0.107]$ & $+0.087$ $[+0.042, +0.140]$ \\
\bottomrule
\end{tabular}
\caption{Cross-scorer robustness ($\lambda=0.12$): CAM-DF minus tuned predict-then-threshold, paired bootstrap mean and 95\% CI, per scorer and dispersion. Boldface marks the key readout: all CIs are strictly positive, and gains are largest for the lowest-AUC scorer.}%
\label{tab:cross}%
\end{table*}

\begin{table*}[t]
\centering
\footnotesize
\setlength{\tabcolsep}{2.5pt}
\begin{tabular}{lccccc}
\toprule
 & & \multicolumn{2}{c}{vs.\ pred.-then-thr.\ (feature-matched)} & \multicolumn{2}{c}{vs.\ score-per-cost threshold} \\
\cmidrule(lr){3-4}\cmidrule(lr){5-6}
Scorer & AUC & $d=0$ & $d=1.0$ & $d=0$ & $d=1.0$ \\
\midrule
deepseek-v3       & 0.835 & $+0.098$ $[+0.057, +0.138]$ & $+0.091$ $[+0.046, +0.139]$ & $+0.099$ $[+0.030, +0.174]$ & $+0.122$ $[+0.045, +0.204]$ \\
qwen-plus (cons.) & 0.894 & $+0.084$ $[+0.031, +0.139]$ & $+0.021$ $[-0.015, +0.064]$ & $+0.080$ $[+0.008, +0.156]$ & $+0.101$ $[+0.033, +0.176]$ \\
qwen-turbo        & 0.906 & $+0.063$ $[+0.031, +0.099]$ & $+0.060$ $[+0.020, +0.103]$ & $+0.048$ $[+0.009, +0.084]$ & $+0.023$ $[-0.022, +0.070]$ \\
qwen-plus         & 0.946 & $+0.053$ $[+0.013, +0.096]$ & $+0.057$ $[+0.017, +0.101]$ & $+0.030$ $[-0.016, +0.072]$ & $+0.040$ $[-0.007, +0.082]$ \\
qwen-max          & 0.953 & $+0.072$ $[+0.036, +0.113]$ & $+0.075$ $[+0.033, +0.123]$ & $+0.062$ $[+0.027, +0.102]$ & $+0.044$ $[-0.010, +0.104]$ \\
\bottomrule
\end{tabular}
\caption{Feature-matched cross-scorer control ($\lambda=0.12$, 30 splits, paired task bootstrap): CAM-DF minus each baseline inside the conservative lite-comparison pipeline, where the predict-then-threshold baseline is trained on the cost-inclusive 16-feature public aggregate state of the main table (removing the feature-set confound of the released score-only cross-scorer pipeline in Table~\ref{tab:cross}, whose baseline cannot observe costs). 9 of 10 prediction-baseline CIs remain strictly positive; the tenth stays positive in mean. Ratio-threshold cells also print the per-scorer support for the main-text claim that score-per-cost intervals exclude zero for the weakest scorer at both dispersions.}%
\label{tab:costaware}%
\end{table*}

\begin{table*}[t]
\centering
\small
\begin{tabular}{lccc}
\toprule
CAM-DF $-$ variant & $d=0$ & $d=1.0$ & $d=1.5$ \\
\midrule
No tool one-hot           & $+0.027$ $[-0.013, +0.065]$ & $+0.028$ $[+0.003, +0.056]$ & $+0.023$ $[+0.003, +0.051]$ \\
\keymetric{Aggregate features only}   & $+0.034$ $[-0.006, +0.072]$ & \keymetric{$+0.051$ $[+0.022, +0.083]$} & \keymetric{$+0.047$ $[+0.019, +0.079]$} \\
Uniform weights           & $+0.041$ $[-0.006, +0.098]$ & $+0.034$ $[-0.014, +0.087]$ & $+0.037$ $[-0.009, +0.091]$ \\
\keymetric{CAM-DF-lite}               & \keymetric{$+0.024$ $[-0.016, +0.062]$} & \keymetric{$+0.027$ $[-0.004, +0.063]$} & \keymetric{$+0.022$ $[-0.008, +0.057]$} \\
\bottomrule
\end{tabular}
\caption{Ablations at $\lambda=0.12$: paired bootstrap mean difference (full CAM-DF minus variant) with 95\% CI (plotted in Figure~\ref{fig:ablation}, plus the lite comparison). Boldface marks the two numbers to read first: removing marginal/cost features hurts most under heterogeneous costs, while CAM-DF-lite stays close to full CAM-DF. ``Aggregate features only'' removes the marginal next-tool block, interactions, and tool indicators (= Aggregate-only DF); ``uniform weights'' removes regret weighting. Two main-text cells quoted outside this grid: at $\lambda=0.05$ the uniform-weight gap is $+0.079$ $[+0.020,+0.149]$, $+0.072$ $[+0.016,+0.136]$, $+0.077$ $[+0.024,+0.142]$ for $d=0,1.0,1.5$; and the no-one-hot variant minus predict-then-threshold stopping is $+0.035$ $[+0.000,+0.067]$ at $d=1.0$ and $+0.045$ $[+0.008,+0.082]$ at $d=1.5$ (paired task bootstrap on the released ablation records).}%
\label{tab:ablation}%
\end{table*}

\begin{figure}[t]
\centering
\includegraphics[width=0.66\columnwidth]{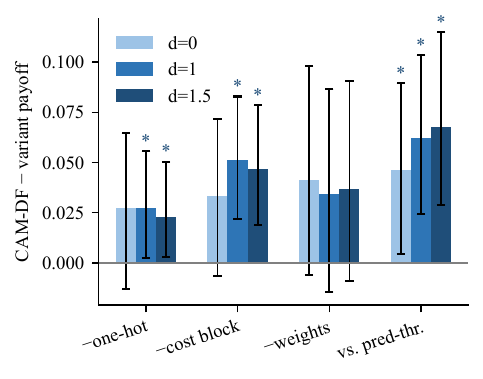}
\caption{Ablations at $\lambda=0.12$ (visualization of Table~\ref{tab:ablation}): paired-bootstrap mean difference (full CAM-DF minus variant) with 95\% CIs, per dispersion $d$. The marginal-cost block matters most under heterogeneous cost; regret weights help directionally at every $d$ (wide CIs at this $\lambda$); the next-tool one-hot costs the least, so the learned-policy gains are not identity memorization. These ablations localize the CAM-DF-family gain. Rightmost group: margin over predict-then-threshold stopping, for scale. Asterisks mark CIs excluding zero.}%
\label{fig:ablation}%
\end{figure}

\begin{table}[t]
\centering
\small
\setlength{\tabcolsep}{2pt}
\begin{tabular}{lcc}
\toprule
$\lambda$ & $d=1.0$ & $d=1.5$ \\
\midrule
0.05 & $-0.003$ $[-0.016, +0.009]$ & $-0.015$ $[-0.043, +0.003]$ \\
0.12 & $+0.003$ $[-0.013, +0.019]$ & $+0.007$ $[-0.029, +0.033]$ \\
0.20 & $+0.005$ $[-0.030, +0.043]$ & \keymetric{$+0.053$ $[+0.002, +0.098]$} \\
\bottomrule
\end{tabular}
\caption{Ordering contrast (qwen-plus scores, paired task bootstrap, 95\% CIs): same-pipeline CAM-DF acquiring in score order minus its $s_j/c_j$ $s_j/c_j$-order variant. At $d=0$ the two orders coincide, so the difference is identically zero (column omitted). No cell significantly favors the $s_j/c_j$ order; at the highest cost pressure and dispersion the score order is significantly better (boldface; task win rate $0.84$).}%
\label{tab:order}%
\end{table}

\begin{table*}[t]
\centering
\small
\setlength{\tabcolsep}{4pt}
\begin{tabular}{llccc}
\toprule
Predicate & $\lambda$ & $d=0$ & $d=1.0$ & $d=1.5$ \\
\midrule
Fixed $0.5$ threshold & 0.05 & $+0.051$ $[-0.008, +0.121]$ & $+0.049$ $[-0.017, +0.118]$ & $+0.054$ $[-0.007, +0.123]$ \\
                      & 0.12 & $+0.022$ $[-0.036, +0.085]$ & $+0.028$ $[-0.033, +0.095]$ & $+0.034$ $[-0.027, +0.099]$ \\
                      & 0.20 & $+0.021$ $[-0.029, +0.077]$ & $+0.048$ $[-0.010, +0.113]$ & $+0.100$ $[+0.034, +0.169]$ \\
\addlinespace
Largest score gap     & 0.05 & $+0.054$ $[-0.005, +0.123]$ & $+0.052$ $[-0.012, +0.125]$ & $+0.058$ $[-0.002, +0.128]$ \\
                      & 0.12 & $+0.030$ $[-0.030, +0.095]$ & $+0.037$ $[-0.024, +0.100]$ & $+0.044$ $[-0.017, +0.112]$ \\
                      & 0.20 & $+0.033$ $[-0.021, +0.092]$ & $+0.062$ $[+0.003, +0.131]$ & $+0.115$ $[+0.049, +0.185]$ \\
\addlinespace
$80\%$ score mass     & 0.05 & $+0.139$ $[+0.059, +0.235]$ & $+0.142$ $[+0.064, +0.228]$ & $+0.149$ $[+0.073, +0.240]$ \\
                      & 0.12 & $+0.110$ $[+0.043, +0.188]$ & $+0.126$ $[+0.054, +0.204]$ & $+0.137$ $[+0.068, +0.212]$ \\
                      & 0.20 & $+0.107$ $[+0.043, +0.175]$ & $+0.151$ $[+0.079, +0.227]$ & $+0.212$ $[+0.134, +0.290]$ \\
\addlinespace
Plug-in $s_j \ge \lambda c_j$ & 0.05 & $+0.048$ $[+0.011, +0.078]$ & $+0.021$ $[-0.017, +0.052]$ & $+0.009$ $[-0.028, +0.039]$ \\
                      & 0.12 & \keymetric{$+0.059$ $[+0.008, +0.103]$} & \keymetric{$+0.071$ $[+0.011, +0.120]$} & \keymetric{$+0.068$ $[+0.007, +0.121]$} \\
                      & 0.20 & $+0.152$ $[+0.088, +0.209]$ & $+0.158$ $[+0.089, +0.222]$ & $+0.143$ $[+0.083, +0.199]$ \\
\bottomrule
\end{tabular}
\caption{Training-free stopping predicates (Retail, qwen-plus scores, paired task bootstrap, 95\% CIs): CAM-DF minus predicate per cost pressure $\lambda$ and dispersion $d$. The predicates carry no tuned or learned parameters. Boldface: the naive $s_j \ge \lambda c_j$ plug-in loses to CAM-DF at the main cost pressure at every dispersion; at $\lambda=0.20$, $d=1.5$ every predicate loses significantly.}%
\label{tab:trainfree}%
\end{table*}

\begin{table*}[t]
\centering
\small
\setlength{\tabcolsep}{3pt}
\begin{tabular}{lccc}
\toprule
Paired difference & $d=0$ & $d=1.0$ & $d=1.5$ \\
\midrule
CAM-DF $-$ CAM-DF (val-tuned)        & $+0.016$ $[-0.006, +0.043]$ & $+0.018$ $[-0.016, +0.059]$ & $+0.007$ $[-0.025, +0.046]$ \\
Pred.\ (tuned) $-$ Pred.\ (fixed 0.5)     & $+0.024$ $[-0.017, +0.074]$ & $+0.015$ $[-0.027, +0.062]$ & $+0.015$ $[-0.026, +0.061]$ \\
CAM-DF $-$ Pred.\ (fixed 0.5)             & $+0.070$ $[+0.015, +0.135]$ & $+0.077$ $[+0.019, +0.135]$ & $+0.083$ $[+0.024, +0.153]$ \\
\keymetric{CAM-DF (val-tuned) $-$ Pred.\ (tuned)} & \keymetric{$+0.030$ $[+0.003, +0.055]$} & \keymetric{$+0.044$ $[+0.009, +0.078]$} & \keymetric{$+0.060$ $[+0.023, +0.095]$} \\
\bottomrule
\end{tabular}
\caption{Tuning-protocol controls ($\lambda=0.12$, qwen-plus scores, paired task bootstrap, 95\% CIs): equalizing the tuning protocol in either direction leaves the CAM-DF gap intact or larger. Boldface marks the protocol-fair headline comparison: validation-tuned CAM-DF still beats tuned predict-then-threshold stopping at all three dispersions. Validation-tuned CAM-DF stop thresholds have per-$d$ medians $0.32$/$0.25$/$0.25$ (19\% fall within $0.5\pm0.1$) without improving payoff; across fixed thresholds $0.3$--$0.5$ the payoff varies by less than $0.01$ at every dispersion.}%
\label{tab:tuning}%
\end{table*}

\begin{table}[t]
\centering
\footnotesize
\setlength{\tabcolsep}{3.2pt}
\begin{tabular}{lccccc}
\toprule
 & $\lambda{=}0.05$ & $0.08$ & $0.12$ & $0.16$ & $0.20$ \\
\midrule
\addlinespace
\multicolumn{6}{l}{\emph{CAM-DF minus predict-then-threshold}}\\
$d=0.0$ & $+0.031^{*}$ & $+0.027$ & $+0.053^{*}$ & $+0.082^{*}$ & $+0.094^{*}$ \\
$d=0.5$ & $+0.028^{*}$ & $+0.040^{*}$ & $+0.060^{*}$ & $+0.077^{*}$ & $+0.076^{*}$ \\
$d=1.0$ & $+0.030^{*}$ & $+0.043^{*}$ & $+0.057^{*}$ & $+0.068^{*}$ & $+0.087^{*}$ \\
$d=1.5$ & $+0.036^{*}$ & $+0.047^{*}$ & $+0.062^{*}$ & $+0.068^{*}$ & $+0.130^{*}$ \\
$d=2.0$ & $+0.038^{*}$ & $+0.050^{*}$ & $+0.074^{*}$ & $+0.091^{*}$ & $+0.208^{*}$ \\
\addlinespace
\multicolumn{6}{l}{\emph{CAM-DF minus score-per-cost threshold}}\\
$d=0.0$ & $+0.018$ & $+0.013$ & $+0.030$ & $+0.043$ & $+0.050^{*}$ \\
$d=0.5$ & $-0.001$ & $+0.016$ & $+0.043$ & $+0.053^{*}$ & $+0.040$ \\
$d=1.0$ & $-0.002$ & $+0.006$ & $+0.040$ & $+0.057^{*}$ & $+0.052$ \\
$d=1.5$ & $-0.005$ & $+0.001$ & $+0.021$ & $+0.041$ & $+0.109^{*}$ \\
$d=2.0$ & $-0.020$ & $-0.025$ & $-0.016$ & $-0.013$ & $+0.095^{*}$ \\
\bottomrule
\end{tabular}
\caption{Dense $5\times5$ comparative-static grid (qwen-plus scores, 30 splits per cell, paired task bootstrap mean; $^{*}$ marks 95\% CIs excluding zero): CAM-DF minus baseline over $\lambda\in\{0.05,\dots,0.20\}$ and dispersion $d\in\{0,\dots,2.0\}$, run through the identical conservative pipeline (the nine main-grid cells reproduce the released values exactly). Against predict-then-threshold stopping, 24 of 25 cells have strictly positive CIs and the gain grows monotonically with $\lambda$ at every $d$ (P3) and with $d$ at high $\lambda$ (P2). Against the tuned score-per-cost threshold, significant cells concentrate at high cost pressure, while at extreme dispersion with low-to-mid pressure ($d=2.0$, $\lambda\le0.16$) the heuristic is directionally ahead (CIs cover zero): the learned gate pays off under pressure, not universally.}%
\label{tab:griddense}%
\end{table}

\begin{table}[t]
\centering
\small
\setlength{\tabcolsep}{3.5pt}
\begin{tabular}{llcccc}
\toprule
$\varepsilon$ & $d$ & CAM-DF & Lite & Pred. & Ratio \\
\midrule
0    & 0   & \textbf{0.400} & 0.378 & 0.361 & 0.376 \\
0    & 1.0 & \textbf{0.331} & 0.307 & 0.273 & 0.293 \\
0.05 & 0   & 0.331 & \textbf{0.360} & 0.331 & 0.353 \\
0.05 & 1.0 & 0.269 & 0.272 & 0.244 & \textbf{0.276} \\
0.1  & 0   & 0.263 & 0.298 & 0.293 & \textbf{0.325} \\
0.1  & 1.0 & 0.228 & 0.237 & 0.225 & \textbf{0.261} \\
0.2  & 0   & 0.174 & 0.183 & 0.196 & \textbf{0.249} \\
0.2  & 1.0 & 0.176 & 0.180 & 0.185 & \textbf{0.213} \\
\bottomrule
\end{tabular}
\caption{Supervision-noise sensitivity ($\lambda=0.12$, 30 splits): test payoff on clean $G_x$ after flipping each tool's train/validation membership in $G_x$ i.i.d.\ with probability $\varepsilon$. Bold: best per row. The learned edge requires $\varepsilon \lesssim 0.05$; by $\varepsilon=0.1$ the tuned score-per-cost threshold dominates, and at $\varepsilon=0.2$ every policy degrades while the single-parameter heuristic degrades least.}%
\label{tab:noise}%
\end{table}

\begin{table*}[!tbp]
\centering
\begin{minipage}{0.48\textwidth}
\centering
\scriptsize
\setlength{\tabcolsep}{2.5pt}
\begin{tabular}{lccc}
\toprule
Feature & $d=0$ & $d=1.0$ & $d=1.5$ \\
\midrule
Intercept                          & $+1.84$ (0.04) & $+1.92$ (0.04) & $+1.98$ (0.04) \\
Progress $t/m$                     & $-0.50$ (0.02) & $-0.72$ (0.03) & $-0.79$ (0.03) \\
Next score                         & $+0.53$ (0.01) & $+0.64$ (0.01) & $+0.70$ (0.01) \\
Next score$/$cost                  & $+0.53$ (0.01) & $+0.24$ (0.02) & $+0.04$ (0.01) \\
Next score gap                     & $+0.36$ (0.03) & $+0.28$ (0.03) & $+0.25$ (0.03) \\
Rem.\ max score                    & $+0.53$ (0.01) & $+0.64$ (0.01) & $+0.70$ (0.01) \\
Rem.\ score (norm.)                & $+0.12$ (0.01) & $+0.29$ (0.01) & $+0.36$ (0.01) \\
Rem.\ score$/$cost                 & $+0.14$ (0.03) & $+0.16$ (0.04) & $+0.12$ (0.04) \\
Selected cost                      & $-0.50$ (0.02) & $-0.34$ (0.02) & $-0.21$ (0.02) \\
$\lambda\times$next cost           & $0.00$ (0.00) & $-0.13$ (0.03) & $-0.33$ (0.03) \\
Next high-cost indicator           & $0.00$ (0.00) & $-0.57$ (0.03) & $-0.58$ (0.03) \\
\bottomrule
\end{tabular}
\caption{CAM-DF-lite coefficients at $\lambda=0.12$: mean (SE) over 30 splits, by cost dispersion $d$. CAM-DF-lite is fit in the \emph{continue} orientation (it models $\Pr(\Delta<0)$ and stops when this probability falls below its tuned threshold; Algorithm~\ref{alg:deploy}), so positive coefficients push toward continuing: score-side features push toward continuing, cost-side features push toward stopping, and cost-side weights activate as dispersion grows.}%
\label{tab:coef}%
\end{minipage}
\end{table*}

The coefficient table ties the robustness checks back to the mechanism claim.
CAM-DF-lite does not need hidden labels or tool identities at deployment time: its dominant directions are the score-side continuation terms and the cost-side stopping terms predicted by the marginal rule.

\subsection{External Validation and Complete Numeric Readouts}

\begin{table*}[!tbp]
\centering
\small
\setlength{\tabcolsep}{3pt}
\begin{minipage}[t]{0.485\textwidth}
\centering
\textbf{Mean payoff}\\[0.4ex]
{\renewcommand{\arraystretch}{1.44}%
\begin{tabular*}{\linewidth}{@{\extracolsep{\fill}}lccc@{}}
\toprule
Policy & $0.06$ & $0.12$ & $0.20$ \\
\midrule
Direct-all       & 0.083 & $-0.835$ & $-2.058$ \\
Fixed-$k$        & 0.474 & 0.202 & 0.024 \\
Score thresh.    & 0.470 & 0.204 & 0.016 \\
Pred.-thr.\       & 0.478 & 0.198 & 0.023 \\
CAM-DF-lite      & \textbf{0.482} & \textbf{0.208} & \textbf{0.036} \\
\midrule
Prefix oracle    & 0.607 & 0.332 & 0.119 \\
\bottomrule
\end{tabular*}}
\end{minipage}
\hfill
\begin{minipage}[t]{0.485\textwidth}
\centering
\textbf{CAM-DF-lite minus baseline, task bootstrap}\\[0.4ex]
{\renewcommand{\arraystretch}{1.04}%
\begin{tabular*}{\linewidth}{@{\extracolsep{\fill}}llccc@{}}
\toprule
Baseline & $\lambda$ & Mean & 95\% CI & Win \\
\midrule
Pred.-thr.\ & 0.06 & $+0.0052$ & $[+0.0010, +0.0097]$ & 0.41 \\
             & 0.12 & $+0.0110$ & $[+0.0066, +0.0155]$ & 0.41 \\
             & 0.20 & $+0.0148$ & $[+0.0083, +0.0216]$ & 0.38 \\
\addlinespace[0.2ex]
Score thresh. & 0.06 & $+0.0142$ & $[+0.0045, +0.0240]$ & 0.41 \\
              & 0.12 & $+0.0056$ & $[-0.0028, +0.0139]$ & 0.40 \\
              & 0.20 & $+0.0234$ & $[+0.0152, +0.0322]$ & 0.48 \\
\addlinespace[0.2ex]
Fixed-$k$     & 0.06 & $+0.0084$ & $[+0.0006, +0.0159]$ & 0.53 \\
              & 0.12 & $+0.0061$ & $[-0.0017, +0.0141]$ & 0.45 \\
              & 0.20 & $+0.0130$ & $[+0.0065, +0.0193]$ & 0.36 \\
\bottomrule
\end{tabular*}}
\end{minipage}
\caption{MCP-Atlas external validation ($n=495$, partial payoff, qwen-plus scores, 20 seeds). Left: mean payoff; SEs are $\le0.006$ for all policies except direct-all ($\le0.011$), and bold marks the best non-oracle. Right: paired task bootstrap over the 485 tasks with complete pairs, reporting CAM-DF-lite minus each baseline with task win rate. Versus direct-all the paired differences are $+0.39$ to $+2.08$ (omitted).}%
\label{tab:atlas}%
\end{table*}

\begin{figure*}[t]
\centering
\includegraphics[width=\textwidth]{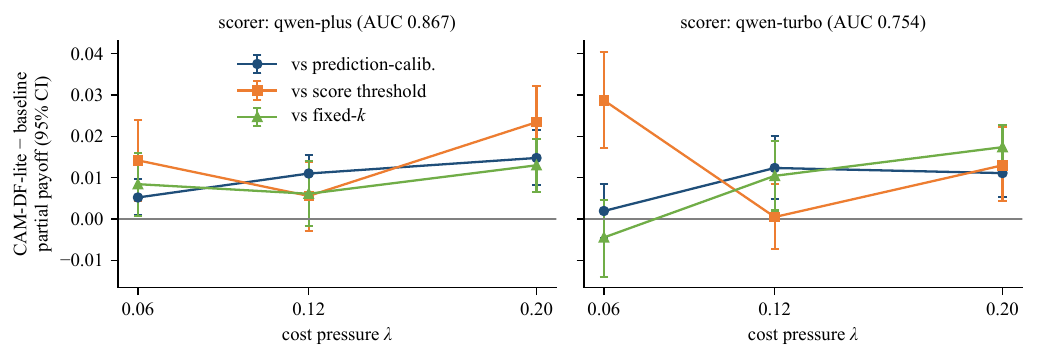}
\caption{MCP-Atlas external validation (partial payoff; paired bootstrap over the 485 of 495 tasks with complete pairs; visualization of Table~\ref{tab:atlas}): task-level paired difference (CAM-DF-lite minus baseline) with 95\% bootstrap CIs at three cost pressures, for the main qwen-plus scorer (left, AUC $0.867$) and a qwen-turbo replication (right, AUC $0.754$). With qwen-plus the difference over predict-then-threshold stopping is positive with CI above zero at all three $\lambda$; over score threshold and fixed-$k$ the CI covers zero only at $\lambda=0.12$. With qwen-turbo, at $\lambda=0.06$ the mean difference versus fixed-$k$ is negative ($-0.005$) and the CI versus predict-then-threshold covers zero; at $\lambda=0.12$ and $0.20$ the CIs versus predict-then-threshold are strictly positive.}%
\label{fig:atlas}%
\end{figure*}

\begin{table*}[!tbp]
\centering
\small
\setlength{\tabcolsep}{5pt}
\begin{tabular}{lccc}
\toprule
CAM-DF-lite $-$ baseline & $\lambda{=}0.06$ & $\lambda{=}0.12$ & $\lambda{=}0.20$ \\
\midrule
Tuned fixed-$k$ & $+0.0169^{*}$ (10/10) & $+0.0183^{*}$ (9/10) & $+0.0431^{*}$ (10/10) \\
Score threshold & $+0.0210^{*}$ (10/10) & $+0.0162^{*}$ (10/10) & $+0.0448^{*}$ (10/10) \\
Score-per-cost threshold & $+0.0091^{*}$ (9/10) & $+0.0056^{*}$ (8/10) & $-0.0106^{\dagger}$ (1/10) \\
Predict-then-threshold & $+0.0093^{*}$ (9/10) & $+0.0295^{*}$ (10/10) & $+0.0655^{*}$ (10/10) \\
Cost-blind lite & $+0.0072^{*}$ (9/10) & $+0.0105^{*}$ (9/10) & $+0.0288^{*}$ (10/10) \\
\bottomrule
\end{tabular}
\caption{MCP-Atlas heterogeneous-cost replay ($n=495$ scored tasks, $439$ entering the paired bootstrap; partial payoff, qwen-plus scores): ten deterministic pseudo-random cost vectors over the 220-tool corpus (base $\in[0.5,2.0]$ from a salted hash of the tool name, mean-normalized, Retail dispersion interpolation at $d=1.0$), 10 seeds per vector. Cells: pooled paired task-bootstrap mean of CAM-DF-lite minus baseline ($^{*}$ = 95\% CI above zero, $^{\dagger}$ = below zero) and the number of the ten cost vectors with a positive mean. Because no author-chosen vector drives the result, this extends the unit-cost external validation of Table~\ref{tab:atlas}: the decision-focused gain over predict-then-threshold stopping and over a cost-blind lite variant (the unit-cost replay's cost-free features) holds at every $\lambda$, while the tuned score-per-cost threshold overtakes CAM-DF-lite at the highest cost pressure.}%
\label{tab:atlashet}%
\end{table*}

\paragraph{Heterogeneous costs on MCP-Atlas: independent replication.}%
\label{app:atlashetrep}%
Independently of the ten-vector replay above (Table~\ref{tab:atlashet}), we rerun the Atlas replay under a second cost-randomization design: per seed, one base cost per distinct tool name is drawn from $\mathrm{Uniform}(0.5,2.0)$ (the range spanned by the Retail base costs) and interpolated with the Retail dispersion rule $c_j(d)=\max\{0.10,\, 1+d(\bar c_j/\bar c_{\mathrm{mean}}-1)\}$, with $\lambda\in\{0.05,0.12,0.20\}$ and 20 seeds.
The cost-augmented features are constant at $d=0$, so the $d=0$ models coincide with the homogeneous-cost pipeline (a built-in sanity check).
Table~\ref{tab:athet} reports the cost-augmented stopping policy against its cost-blind lite twin and the tuned score-per-cost threshold.
Against the cost-blind twin, all six heterogeneous partial-payoff cells are positive with CIs excluding zero, growing with dispersion; under exact payoff the tuned score-per-cost threshold flips from ahead at $d=0$ to behind at $d=1.5$ ($+0.034$ $[+0.022, +0.046]$ at $\lambda=0.12$; $+0.073$ $[+0.068, +0.079]$ at $\lambda=0.20$).
The honest boundary is the partial-payoff, high-dispersion corner, where the non-prefix score-per-cost rule wins ($-0.042$ $[-0.049, -0.035]$ at $\lambda=0.20$, $d=1.5$): partial credit rewards skipping expensive mid-ranking tools, which a prefix policy cannot do.
Prefix stopping is robust under exact sufficiency; under partial credit with extreme heterogeneity, non-prefix marginal rules become the stronger competitor, again evidence that the marginal value/cost structure, rather than the specific policy class, carries the gain.

\begin{table*}[!tbp]
\centering
\small
\setlength{\tabcolsep}{4pt}
\begin{tabular}{llccc}
\toprule
Comparison & $\lambda$ & $d=0$ & $d=1.0$ & $d=1.5$ \\
\midrule
Partial: minus cost-blind lite & 0.05 & $+0.000$ & $+0.008$ $[+0.004, +0.012]$ & $+0.012$ $[+0.007, +0.016]$ \\
                             & 0.12 & $+0.000$ & $+0.020$ $[+0.015, +0.026]$ & \keymetric{$+0.039$ $[+0.033, +0.046]$} \\
                             & 0.20 & $+0.000$ & $+0.033$ $[+0.028, +0.039]$ & \keymetric{$+0.063$ $[+0.057, +0.070]$} \\
\addlinespace
Exact: minus score-per-cost threshold & 0.05 & $-0.024$ $[-0.047, -0.000]$ & $+0.002$ $[-0.018, +0.024]$ & $+0.001$ $[-0.018, +0.021]$ \\
                             & 0.12 & $-0.011$ $[-0.015, -0.006]$ & $+0.023$ $[+0.013, +0.032]$ & $+0.034$ $[+0.022, +0.046]$ \\
                             & 0.20 & $-0.001$ $[-0.002, -0.000]$ & $+0.057$ $[+0.053, +0.061]$ & $+0.073$ $[+0.068, +0.079]$ \\
\addlinespace
Partial: minus score-per-cost threshold & 0.05 & $+0.016$ $[+0.007, +0.025]$ & $+0.008$ $[-0.000, +0.016]$ & $+0.002$ $[-0.005, +0.009]$ \\
                             & 0.12 & $+0.006$ $[-0.002, +0.014]$ & $+0.010$ $[+0.004, +0.017]$ & $-0.009$ $[-0.016, -0.002]$ \\
                             & 0.20 & $+0.021$ $[+0.014, +0.029]$ & $-0.005$ $[-0.011, +0.002]$ & \keymetric{$-0.042$ $[-0.049, -0.035]$} \\
\bottomrule
\end{tabular}
\caption{MCP-Atlas heterogeneous-cost replay (random $\mathrm{Uniform}(0.5,2.0)$ base costs per seed, Retail dispersion interpolation, 20 seeds; paired task bootstrap over 495 tasks, 95\% CIs): cost-augmented stopping policy minus baseline. Top: against the identical policy without cost features (identically zero at $d=0$ by construction); all six heterogeneous cells are positive and grow with dispersion (boldface: largest). Middle: against the tuned score-per-cost threshold under exact payoff: behind at $d=0$, ahead at $d\ge1.0$ for $\lambda\ge0.12$. Bottom: the same comparison under partial payoff: at the highest pressure and dispersion the non-prefix score-per-cost rule wins (boldface), because partial credit rewards skipping expensive mid-ranking tools that a prefix policy must pass through.}%
\label{tab:athet}%
\end{table*}

\paragraph{MetaTool reduced boundary check.}
MetaTool \citep{huang2024metatool} is a useful stress test for whether a query-to-tool label benchmark can support our acquisition abstraction, but it is not a native cost-aware stopping benchmark: the public multi-tool subset gives two required tools per query, not live tool outputs, operational costs, or a downstream task reward.
We therefore ran a no-API reduced replay on the 497-task corpus with the 15 tools appearing as required in that subset, unit costs, exact sufficiency, and an unsupervised train-split TF-IDF cosine scorer (full-fit diagnostic AUC $0.765$, mean cover-all rank $6.82$).
The 30 random 60/20/20 splits place 100 tasks in each test fold and cover 496 unique tasks in the paired bootstrap table.
The result is deliberately not used as main evidence.
At $\lambda=0.12$, CAM-DF-lite beats predict-then-threshold stopping by $+0.026$ $[+0.002,+0.050]$, but is not better than fixed-$k$ ($-0.009$ $[-0.025,+0.005]$) or score threshold ($-0.014$ $[-0.031,+0.002]$).
This check supports the boundary stated in the main paper: CAM-DF needs a meaningful ranking signal and payoff/cost labels; shallow homogeneous-cost tool-selection labels are not enough to claim a third positive benchmark.

\begin{table*}[!tbp]
\centering
\footnotesize
\setlength{\tabcolsep}{3.5pt}
\begin{tabular}{llccc}
\toprule
CAM-DF $-$ baseline & $\lambda$ & $d=0$ & $d=1.0$ & $d=1.5$ \\
\midrule
Tuned fixed-$k$ & 0.05 & $+0.048$ $[+0.019, +0.075]$ (.93) & $+0.032$ $[-0.001, +0.060]$ (.91) & $+0.029$ $[-0.002, +0.055]$ (.85) \\
 & 0.12 & $+0.126$ $[+0.076, +0.174]$ (.81) & $+0.111$ $[+0.065, +0.157]$ (.82) & $+0.100$ $[+0.053, +0.146]$ (.82) \\
 & 0.20 & $+0.087$ $[+0.028, +0.140]$ (.76) & $-0.018$ $[-0.091, +0.051]$ (.33) & $-0.038$ $[-0.112, +0.030]$ (.34) \\
\addlinespace
Score threshold & 0.05 & $+0.018$ $[-0.014, +0.054]$ (.51) & $+0.012$ $[-0.028, +0.052]$ (.51) & $+0.008$ $[-0.027, +0.045]$ (.54) \\
 & 0.12 & $+0.030$ $[-0.013, +0.071]$ (.45) & $+0.032$ $[-0.013, +0.077]$ (.55) & $+0.035$ $[-0.014, +0.081]$ (.54) \\
 & 0.20 & $+0.053$ $[+0.010, +0.101]$ (.43) & $+0.088$ $[+0.042, +0.145]$ (.57) & $+0.147$ $[+0.094, +0.207]$ (.73) \\
\addlinespace
Score-per-cost threshold & 0.05 & $+0.018$ $[-0.013, +0.054]$ (.51) & $-0.002$ $[-0.038, +0.038]$ (.48) & $-0.005$ $[-0.044, +0.034]$ (.51) \\
 & 0.12 & $+0.030$ $[-0.010, +0.073]$ (.45) & $+0.040$ $[-0.005, +0.083]$ (.63) & $+0.021$ $[-0.031, +0.067]$ (.73) \\
 & 0.20 & $+0.050$ $[+0.008, +0.094]$ (.57) & $+0.052$ $[-0.013, +0.114]$ (.58) & $+0.109$ $[+0.050, +0.161]$ (.88) \\
\addlinespace
Retail rule & 0.05 & $+0.053$ $[+0.000, +0.112]$ (.39) & $+0.042$ $[-0.014, +0.105]$ (.34) & $+0.042$ $[-0.016, +0.105]$ (.36) \\
 & 0.12 & $+0.054$ $[-0.009, +0.119]$ (.46) & $+0.037$ $[-0.026, +0.102]$ (.45) & $+0.031$ $[-0.030, +0.097]$ (.43) \\
 & 0.20 & $+0.086$ $[+0.022, +0.147]$ (.54) & $+0.081$ $[+0.018, +0.145]$ (.57) & $+0.124$ $[+0.053, +0.189]$ (.64) \\
\addlinespace
Predict-then-threshold & 0.05 & $+0.031$ $[+0.003, +0.064]$ (.33) & $+0.027$ $[-0.007, +0.064]$ (.36) & $+0.032$ $[+0.000, +0.068]$ (.37) \\
 & 0.12 & $+0.046$ $[+0.005, +0.089]$ (.52) & $+0.062$ $[+0.024, +0.104]$ (.54) & $+0.068$ $[+0.026, +0.112]$ (.46) \\
 & 0.20 & $+0.086$ $[+0.042, +0.135]$ (.49) & $+0.113$ $[+0.064, +0.167]$ (.60) & $+0.167$ $[+0.110, +0.227]$ (.76) \\
\addlinespace
Aggregate-only DF & 0.05 & $+0.014$ $[+0.000, +0.026]$ (.82) & $+0.005$ $[-0.021, +0.023]$ (.85) & $+0.002$ $[-0.026, +0.021]$ (.82) \\
 & 0.12 & $+0.034$ $[-0.008, +0.071]$ (.60) & $+0.051$ $[+0.023, +0.083]$ (.60) & $+0.047$ $[+0.019, +0.081]$ (.54) \\
 & 0.20 & $+0.047$ $[+0.006, +0.084]$ (.37) & $+0.074$ $[+0.039, +0.112]$ (.57) & $+0.092$ $[+0.058, +0.124]$ (.73) \\
\bottomrule
\end{tabular}
\caption{Full $\lambda$-grid Retail task-level paired bootstrap (qwen-plus scores, 67 tasks, 2{,}000 resamples): mean payoff difference (CAM-DF minus baseline), 95\% CI, and task win rate in parentheses, for every $\lambda \in \{0.05, 0.12, 0.20\}$ and dispersion $d$. The $\lambda=0.12$ rows are the main comparison quoted in Section~\ref{sec:exp-main}.}%
\label{tab:grid}%
\end{table*}

\end{document}